\documentclass[10pt,twocolumn,letterpaper]{article}

\usepackage[pagenumbers]{cvpr} 

\definecolor{cvprblue}{rgb}{0.21,0.49,0.74}
\usepackage[pagebackref,breaklinks,colorlinks,allcolors=cvprblue]{hyperref}

\usepackage{multirow}
\usepackage{algorithm}
\usepackage{algorithmic}
\usepackage{booktabs}
\usepackage{multirow}
\usepackage{svg}

\usepackage{booktabs}
\usepackage{multirow}
\usepackage{siunitx}
\usepackage{tabularx}
\usepackage{adjustbox}
\newcolumntype{Y}{>{\centering\arraybackslash}X}
\usepackage{makecell}

\usepackage{cellspace}
\setcellgapes{6pt}

\usepackage{caption}
\usepackage{booktabs} 
\usepackage{makecell} 
\usepackage{multirow} 
\usepackage{amssymb} 
\usepackage{caption} 
\usepackage{amsmath} 
\usepackage{amssymb} 
\usepackage{bm}      

\newcommand{\fn}{{\mathrm{MLP}}} 
\newcommand{\linear}{{\mathrm{Linear}}} 
\usepackage[capitalize]{cleveref}

\usepackage{tcolorbox}
\usepackage{multibib}

\newcites{app}{Appendix References}  
\usepackage{newfloat}
\usepackage{listings}
\floatstyle{ruled}
\newfloat{listing}{tb}{lst}{}
\floatname{listing}{Listing}

\def\paperID{29850} 
\def\confName{CVPR}
\def\confYear{2026}

\title{MeteorPred: A Meteorological Multimodal Large Model and Dataset for Severe Weather Event Prediction}

\author{
Shuo Tang$^{1,2,3}$,
Jian Xu$^{1,2\dagger}$,
Jiadong Zhang$^{1,2,3}$,
Yi Chen$^{1,2,3}$,
Qizhao Jin$^{4}$,
Lingdong Shen$^{1,2}$,\\
Chenglin Liu$^{1,2,3}$,
Shiming Xiang$^{1,2,3}$\\[2pt]
$^{1}$MAIS, Institute of Automation, Chinese Academy of Sciences\\
$^{2}$School of Artificial Intelligence, University of Chinese Academy of Sciences\\
$^{3}$Zhongguancun Academy, Beijing\\
$^{4}$China Meteorological Administration\\[4pt]
\tt\small \{tangshuo2024, jian.xu\}@ia.ac.cn}

\begin{document}
\maketitle
\begin{abstract}

\vspace{-0.8cm}
Timely and accurate forecasts of severe weather events are essential for early warning and for constraining downstream analysis and decision-making. Since severe weather events prediction still depends on subjective, time-consuming expert interpretation, end-to-end “AI weather station” systems are emerging but face three major challenges: (1) scarcity of severe weather event samples; (2) imperfect alignment between high-dimensional meteorological data and textual warnings; (3) current multimodal language models cannot effectively process high-dimensional meteorological inputs or capture their complex spatiotemporal dependencies. To address these challenges, we introduce MP-Bench, the first large-scale multimodal dataset for severe weather events prediction, comprising 421,363 pairs of raw multi-year meteorological data and corresponding  text caption,  covering a wide range of severe weather scenarios. On top of this dataset, we develop a Meteorology Multimodal Large Model (MMLM) that directly ingests 4D meteorological inputs. In addition, it is designed to accommodate the unique characteristics of 4D meteorological data flow, incorporating three plug-and-play adaptive fusion modules that enable dynamic feature extraction and integration across temporal sequences, vertical pressure layers, and spatial dimensions. Extensive experiments on MP-Bench show that MMLM achieves strong performance across multiple tasks, demonstrating effective severe weather understanding and representing a key step toward automated, AI-driven severe weather events forecasting systems. Our source code and dataset will be made publicly available.
\end{abstract}

\begin{figure}[!h]       
	\centering
	\includegraphics[width=\columnwidth]{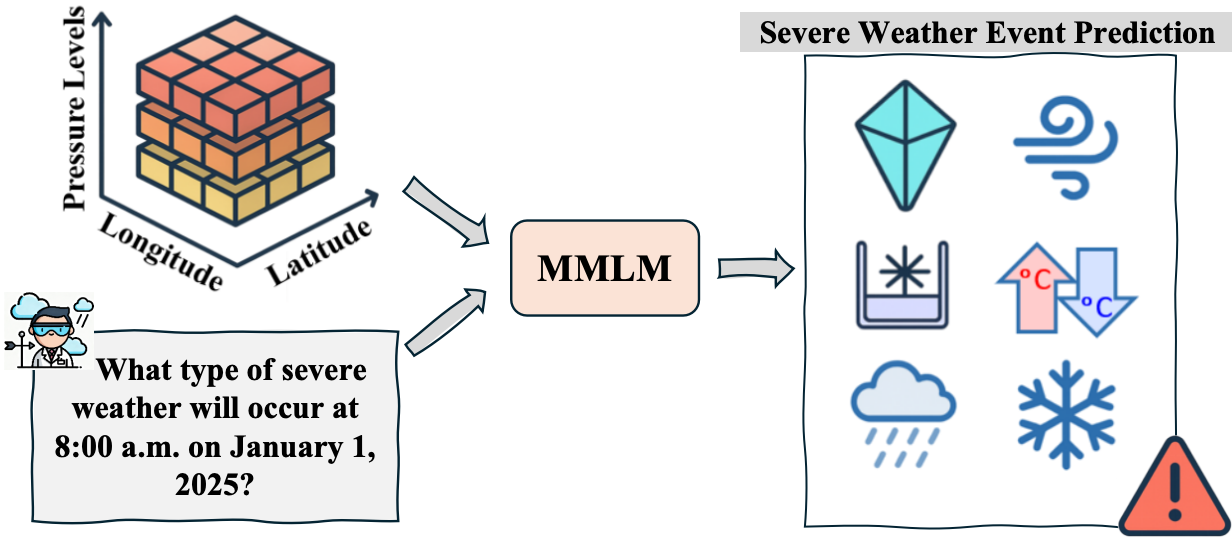}
	\caption{Conceptual illustration for severe weather event prediction using the Meteorological Multimodal Large Model (MMLM).}
	\label{fig:concept}
	\vspace{-0.5cm} 
\end{figure}

\section{Introduction}
Severe weather events, including rain storm, snow storm, hail, gale, frost, heat wave, cold wave, and other high-impact atmospheric phenomena, are occurring with increasing frequency \cite{hoeppe2016trends,stephenson2008definition}. These hazards impose escalating strains worldwide on transportation systems \cite{liu2023}, energy infrastructure \cite{perera2020}, agricultural production \cite{lesk2016}, and public safety \cite{liao2025}. Consequently, timely and accurate forecasts are critical for effective emergency response and disaster mitigation, supporting vital decision-making in public safety, transportation, and industrial production.

\begin{figure*}[t]               
  \centering
  \includegraphics[
    width=\textwidth,          
    height=0.30\textheight,     
    keepaspectratio            
  ]{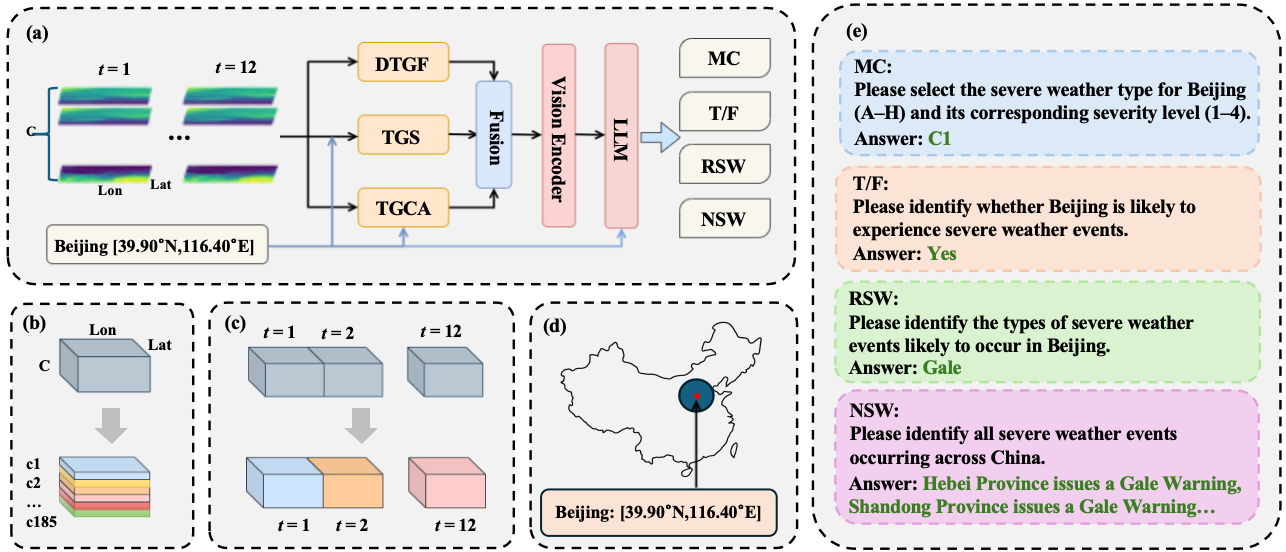}
  \caption{Overview of the MMLM framework and its core components. (a) Displays the MMLM architecture, where outputs from the DTGF, TGCA, and TGS modules are concatenated and integrated in a learnable Fusion Layer before being fed into the LLM. (b), (c), and (d) illustrate the three plug-and-play modules, where color intensity represents the adaptive weights across temporal, channel, and spatial dimensions. (e) shows four QA samples from MP-Bench.}
  \label{fig:enter-label1}
  \vspace{-0.1cm}
\end{figure*}

The meteorological variables in severe weather are complex and variable in their spatial and temporal distribution. Currently, severe weather warnings have to follow a rigorous process. First, the source data observed from satellites, radar and ground stations are assimilated into Numerical Weather Prediction (NWP) systems \cite{bauer2015} and modern AI-based forecasting models for weather prediction \cite{chen2023,price2025}. Then, operational forecasters analyze comprehensively the output of the models and correct the details to formally release the warnings to the public. However, manually interpreting, drafting, and reviewing forecasts is not only time-consuming but also heavily dependent on the forecaster's expertise and subjective judgment, inevitably increasing the risk of oversight \cite{AMS2021WeatherAnalysis}. This challenge makes the development of an AI-driven severe weather warning system essential---a system that ingests the latest NWP output, instantly produces accurate warnings, and provides continuously rolling updates, representing the future of severe weather forecasting through a fully automated early-warning pipeline \cite{reichstein2025early}.

The development of end-to-end AI-based severe weather systems (as shown in \cref{fig:concept}) is currently hindered by three major challenges. First, existing severe weather datasets are often small-scale, event-specific, or geographically and temporally limited, making it difficult to train and evaluate models with strong generalization capabilities \cite{kafi2024advances,olivetti2024advances}. Second, the alignment between high-dimensional meteorological data and textual descriptions of severe weather events remains imperfect, and compressing meteorological data into daily averages further reduces temporal resolution, impairing the ability to capture rapid or dramatic atmospheric changes on short time scales. Third, there are no existing multimodal large language models (MLLMs) that can handle the raw meteorological data well. Consequently, for the gridded data of vertical pressure layers obtained at each moment, most previous methods manually select a subset to adapt to the input requirements of existing encoders. This over-simplification discards critical information, such as vertical structure, temporal dynamics, and inter-variable physical relationships, leading to a significant degradation in prediction accuracy. Moreover, 4D gridded data are often flattened or projected into two-dimensional visual formats, which hinders MLLMs from capturing the intrinsic physical dependencies and spatiotemporal evolution of atmospheric systems.

To address these challenges, we present a unified framework consisting of a large-scale benchmark dataset and a meteorology multimodal large model. Specifically, we constructed a brand new dataset, MP-Bench, which is a nationwide, year-round coverage severe weather events dataset. It contains 421,363 data pairs, including rain storm, snow storm, hail, gale, frost, heat wave, cold wave and normal weather samples. Using each warning’s issuance time as the reference point, we collect the national multi-variable meteorological fields for the next 12 hours and pair them with that the warning to ensure precise cross-modal temporal alignment. In order to fully demonstrate the language comprehension capability and application potential of MLLM, we constructed four types of QA pairs, including \textbf{Multiple Choice Questions}, \textbf{True/False Questions}, \textbf{Regional Severe Weather Questions} and \textbf{National Severe Weather Questions}. With this work setting, it is hoped to enrich the downstream task scenarios.

On this data foundation, we develop the MMLM, a multimodal model tailored for 4D meteorological data. The model integrates three plug-and-play modules: Dynamic Temporal Gating Fusion (DTGF), Text-Driven Gaussian Spatial Masking (TGS), and Text-Driven Channel Attention (TGCA). These modules enhance feature extraction along temporal, spatial, and vertical dimensions, respectively. The synergistic interaction among these modules significantly improves MMLM's capacity to capture and interpret complex multidimensional meteorological patterns. 

Our contributions could be summarized as:

\begin{itemize}
\item We have collected MP-Bench, a dataset for severe weather events prediction that provides nationwide scope, year-round coverage, and a rich variety of Q\&A formats.
\item Based on this dataset, we proposed MMLM, integrating three plug-and-play modules that enhance feature extraction in temporal, spatial, and vertical dimensions, respectively, collectively improving its ability to capture complex meteorological patterns.

\begin{table*}[!htbp]
\centering
\small
\begin{tabular}{lccccc}
\toprule
Dataset & Meteorological Variables & Text Events & Severe Weather Types & Temporal Alignment  \\
\midrule
CrisisLex \cite{olteanu2014crisislex} & \texttimes & 2,840,000 & 2 & \texttimes  \\
ClimaBench \cite{laud2023climabench} & \texttimes & 37,989 & 6 & \texttimes  \\
Climate-FEVER \cite{diggelmann2020climatefever} & \texttimes & 1,535 & 5 & \texttimes  \\
GridRad-Severe \cite{DevelopmentandInvestigationofGridRadSevereaMultiyearSevereEventRadarDataset} & Remote Sensing & \texttimes & 3 & \texttimes \\
SEVIR \cite{veillette2020sevir} & Remote Sensing & \texttimes & 5 & \texttimes  \\
HR-Extreme \cite{ran2024hr} & NWP & \texttimes & 5 & \texttimes \\
WeatherQA \cite{ma2024weatherqa} & Reanalysis, Observation & 8,000 & 5 & 1h\\ 
ClimateIQA \cite{chen2024vision} & Reanalysis & 762,120 & 1 & 1h\\
CLLMate \cite{li2024cllmate} & Reanalysis & 41,000 & 3 & 1d mean\\
OmniEarth-Bench \cite{wang2025omniearth} & Remote Sensing & 6,395 & 5 & 4h (1h interval)/1d (6h interval) \\
\textbf{MP-Bench} & \textbf{Reanalysis} & \textbf{421,363} & \textbf{7} & \textbf{12h (1h interval)}  \\
\bottomrule
\end{tabular}
\vspace{-0.1cm}
\captionof{table}{Comparison between MP-Bench and existing multimodal datasets for weather- and disaster-related analysis. Here we summarize the type of meteorological variables used, the scale of associated text events, the number of severe weather types, and the temporal alignment strategy between meteorological fields and textual events. 1h=1 hour, 1d=1 day.}
\label{tab:extreme_weather_benchmarks}
\vspace{-0.3cm}
\end{table*}

\item To the best of our knowledge, this is the first time that MLLM has been used to deeply interpret raw meteorological data and generate warning conclusions in sentence form, paving a whole new perspective for future severe weather warning missions.
\end{itemize}

\section{Related Works}
\subsection{Severe Weather Event Prediction}
Contemporary methodologies for AI-based prediction of severe weather events can be conventionally classified into three principal paradigms: foundational models developed on  gridded meteorological data, intelligent inference approaches leveraging LLMs, and end-to-end forecasting frameworks grounded on MLLMs that assimilate heterogeneous data sources.

There are well-established studies leveraging gridded meteorological data to drive fundamental models \cite{lam2023learning,bi2023accurate,ravuri2021skilful,wu2023interpretable,xiao2023towards,gan2025ewmoe}, which directly predict spatial distributions of meteorological variables at the next time step. However, they lack high-level semantic representations of discrete severe-weather phenomena, limiting their ability to produce intuitive and actionable warnings.

Some studies employ LLMs with retrieval-augmented generation (RAG) frameworks \cite{martelo2024towards,wang2025remflow,li2025save,wang2025omniearth}, compressing high-dimensional meteorological data into daily means via spatial and temporal averaging while extracting structured semantics for expert-level analysis. However, this approach overlooks the physical laws governing meteorological field evolution, limiting its ability to capture complex dynamics and making LLMs prone to hallucinations.

The rapid development of MLLMs has opened up new research perspectives for severe weather event prediction \cite{chen2024vision,ma2024weatherqa,li2024cllmate,wang2025omniearth,bodnar2025}. These studies encode high-dimensional meteorological or environmental data into three channels and incorporate corresponding textual information as input to MLLMs, enabling visual question answering and complex reasoning over severe weather or environmental events. However, such methods still have limitations on data processing. To align meteorological data with textual event records, complex meteorological fields are often compressed into daily averages, resulting in coarse temporal resolution and impairing the ability to capture rapid or dramatic atmospheric changes on short time scales. Meanwhile, to fit the input requirements of  MLLMs, only variables from a few pressure levels are selected and temporally averaged, then compressed into RGB images, a process that may discard critical spatiotemporal information.

\subsection{Related Datasets}
From the comparison (As shown in \cref{tab:extreme_weather_benchmarks}), these datasets are leveraged to develop more robust language models for severe weather understanding. CrisisLex \cite{olteanu2014crisislex}, ClimaBench \cite{laud2023climabench}, and Climate-FEVER \cite{diggelmann2020climatefever} each address language understanding tasks centered on severe weather, including information filtering, reasoning, and scientific claim verification. However, all three datasets rely solely on textual data, and the lack of integration with meteorological modalities limits their ability to capture the full meteorological context and physical grounding of severe weather events.

SEVIR \cite{veillette2020sevir} and GridRad-Severe \cite{DevelopmentandInvestigationofGridRadSevereaMultiyearSevereEventRadarDataset} are datasets derived from remote sensing data, specifically designed to facilitate the detection, classification, and predictive modeling of severe weather events. HR-Extreme \cite{ran2024hr} leverages high-resolution numerical weather prediction data to support research on the spatiotemporal detection and classification of diverse severe weather phenomena. However, relying solely on meteorological data for severe weather forecasting has significant limitations. These datasets often lack semantic interpretability, making them difficult to use directly for issuing warnings.

OmniEarth-Bench \cite{wang2025omniearth}, ClimateIQA \cite{chen2024vision}, WeatherQA \cite{ma2024weatherqa} and CLLMate \cite{li2024cllmate} release multimodal datasets focused on severe weather phenomena and have demonstrated efficacy in severe weather event prediction tasks. However, no existing datasets concurrently provide large-scale textual warnings with comprehensive coverage across diverse severe weather types.

We present MP-Bench, a large-scale multimodal dataset built upon years of severe weather event forecasts. Unlike prior works, it integrates 421,363 textual warnings accumulated nationwide from meteorological stations over multiple years, covering seven typical types of severe weather events while preserving the original temporal information of meteorological data. This establishes a robust foundation for severe weather prediction with large-scale data, broad category coverage, and meteorological feature integrity.

\section{MP-Bench}
\subsection{Dataset Overview}
\subsubsection{Data Source}
The dataset used in this study comprises two components: gridded meteorological data (ERA5) \cite{hersbach2020era5} and a textual dataset of severe weather events. ERA5, developed by the European Centre for Medium-Range Weather Forecasts (ECMWF), is a global atmospheric reanalysis dataset. It provides a comprehensive range of meteorological variables with hourly temporal resolution, covering both land and ocean regions globally at a spatial resolution of 0.25° in latitude and longitude. It also includes 37 vertical pressure levels from 1000hPa to 1hPa, with the specific levels listed in the appendix, enabling a detailed representation of atmospheric thermal and dynamic processes from the troposphere to the top of the stratosphere. In this study, we focus on the region of China. The selected variables include temperature, humidity, precipitation, wind speed, and pressure. Each variable is represented across all 37 vertical pressure levels, which are treated as separate channels in our data structure. The data are sampled at multiple time steps, resulting in a four-dimensional format (time × pressure level × longitude × latitude).

\begin{figure}[!t]       
	\centering
	\includegraphics[width=0.60\columnwidth]{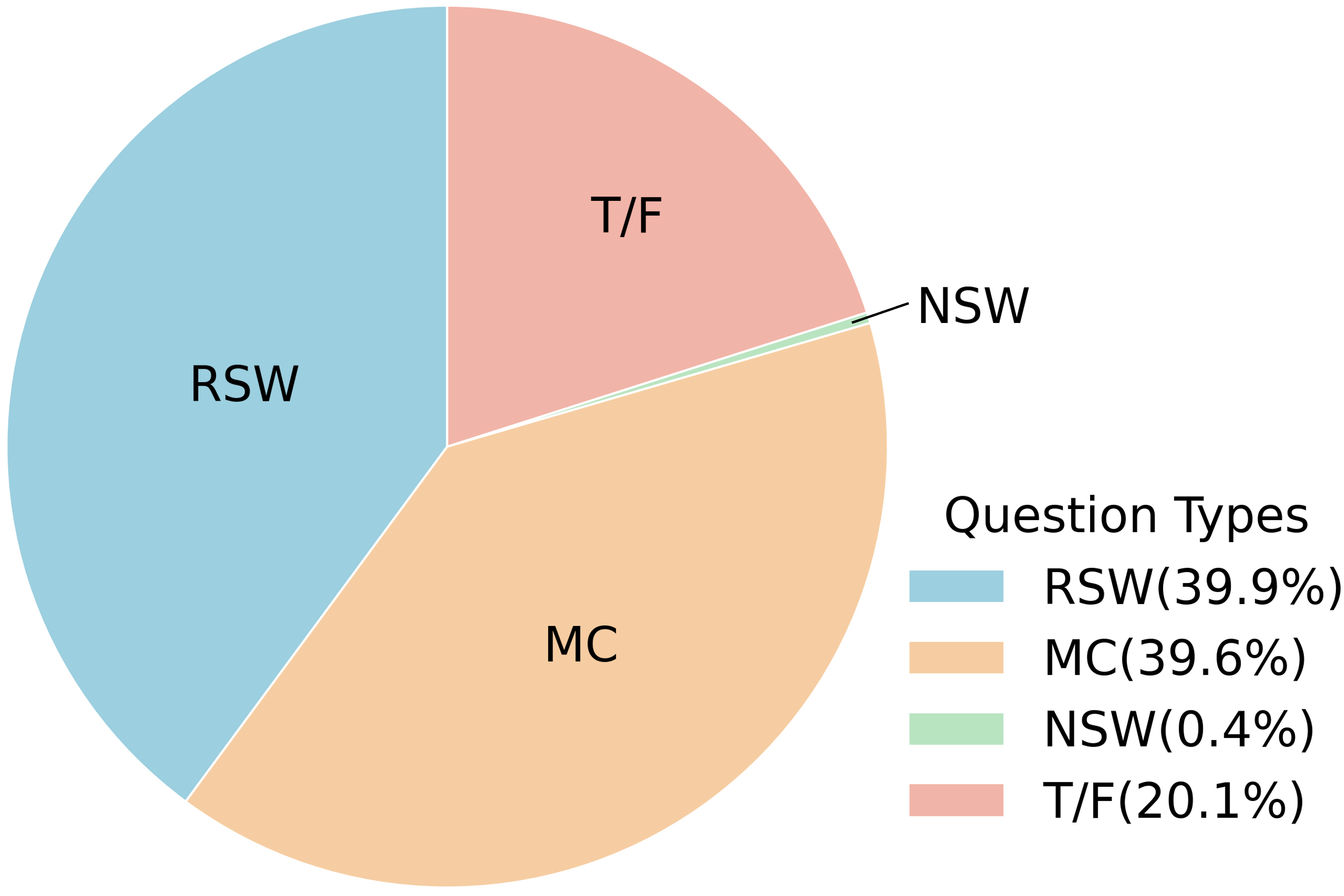}
	\caption{Distribution of four QA task types in MP-Bench, including MC, T/F, RSW, NSW.}
	\label{fig:QA_pie}
	\vspace{-0.5cm} 
\end{figure}

The severe weather event text data were obtained from the China Meteorological Administration (CMA), which provides daily updates of severe weather warnings issued by regional meteorological stations nationwide. The data include records from 2023 and 2024, covering 2,412 weather stations across China. To focus on the most prevalent severe weather types, we filtered the dataset to include seven representative categories for further analysis: rainstorm, snowstorm, gale, cold wave, heat wave, frost, and hail. To maintain labeling consistency, only the issuance time, location, event type, and severity level were retained. Each record indicates that the specified type of severe weather was expected to occur in the associated region within the next few hours. During cleaning, warnings of the same type and station issued within two hours were merged into a single event at the highest severity level. After cleaning, 371,703 valid warning records were obtained. Detailed spatial and weather type distributions are provided in the appendix. To alleviate category imbalance and enhance model discriminative capability, we sampled “normal weather” entries evenly across regions and seasons, adding 49,660 negative samples. In total, the dataset contains 421,363 entries.

Additionally, to comprehensively validate the model’s geographic generalization ability, we also selected a representative subset of the NOAA Storm Events Dataset \cite{ncei_noaa_storm_events_2024}, which records global severe weather events, as an additional dataset for cross-regional generalization verification.

\subsubsection{Data Alignment}
Accurately aligning high-dimensional gridded meteorological data with warning issuance times was challenging. To preserve the original temporal and 3D spatial information in ERA5, we deviated from previous approaches by avoiding temporal averaging. Instead, we extracted a 12-hour window spanning [$t$, $t$+11] hours starting at the warning issuance time $t$, enabling precise alignment of meteorological data with textual events on the time axis.

\subsubsection{QA pairs}
To better fulfill practical application requirements, we designed four distinct types of QA tasks. \cref{fig:QA_pie} illustrates the proportional distribution of these task categories. These task categories differ in terms of spatial granularity, ranging from regional levels (e.g., provinces, cities, and counties) to the national scale, as well as in the level of detail required for querying severe weather events—specifically, whether the query pertains solely to the type of weather event or includes both the type and its severity level. We employ the 2023 data as the training set, and data from 2024 as the test set. Detailed quantity and type distribution of 2023 and 2024 data are provided in the appendix.

\textbf{Multiple Choice Questions (MC)}: These questions are specifically designed to enable fine-grained identification of severe weather events in a given region. Based on the classification criteria for severe weather issued CMA, we categorize severe weather into seven primary types (A–G), with an additional option representing normal weather conditions (H). Each primary category is further subdivided into two to four subcategories based on the severity level of the event (e.g., A.1 denotes a blue rainstorm warning). Each question includes multiple alternative options, from which the model is required to select the single answer that best aligns with the provided meteorological data and the query.

\textbf{True/False Questions (T/F)}: These questions are used to determine whether severe weather is occurring in a specified region, with “Yes” or “No” as the possible responses. The model analyzes the gridded meteorological data and integrates it with the user-defined geographic area to assess whether the region is under severe weather conditions during the target time period.

\begin{table*}[!htbp]
\centering
\begin{tabular}{@{}l l cc cc ccc@{}}
\toprule
\textbf{Category} & \textbf{Model} & \multicolumn{2}{c}{\textbf{MC-main}} & \multicolumn{2}{c}{\textbf{MC-sub}} & \textbf{T/F} & \textbf{RSW} & \textbf{NSW} \\
\cmidrule(r){3-9}
 & & \textbf{Acc ↑} & \textbf{Macro-F1 ↑} & \textbf{Acc ↑} & \textbf{Macro-F1 ↑} & \textbf{Acc ↑} & \textbf{Acc ↑} & \textbf{Score ↑} \\
\midrule
Closed-source model & GPT-4o & 11.92 & 2.92 & 6.51 & 0.88 & 0.19 & 14.03 & 0.1 \\
\midrule
\multirow{4}{*}{Baseline} 
 & LLaVA-NeXT-Video-7B & 47.99 & 20.70 & 38.29 & 8.08 & \textbf{69.23} & 45.22 & 0.2 \\ 
 & Video-LLaVA-7B & \textbf{59.77} & 22.45 & \textbf{49.01} & 8.78 & 68.37 & \textbf{65.31} & 0.6 \\
 & InternVL3-8B & 42.58 & 15.47 & 31.05 & 6.25 & 68.98 & 59.22 & 0.3 \\
 & Qwen2.5-VL-7B-Instruct & 56.26 & \textbf{26.88} & 46.54 & \textbf{9.78} & 68.33 & 61.82 & \textbf{1.7} \\
\midrule
\multirow{4}{*}{MMLM} 
 & LLaVA-NeXT-Video-7B & 55.27 & 23.81 & 43.55 & 9.55 & 79.37 & 53.64 & 0.7 \\
 & Video-LLaVA-7B & 64.46 & 35.73 & 53.49 & 15.88 & 78.65 & 65.81 & 1.7 \\ 
 & InternVL3-8B & 58.31 & 24.18 & 47.03 & 9.79 & 79.40 & 70.03 & 1.9 \\ 
 & Qwen2.5-VL-7B-Instruct & \textbf{72.37} & \textbf{50.88} & \textbf{58.19} & \textbf{29.31} & \textbf{87.13} & \textbf{71.23} & \textbf{2.1} \\
\bottomrule
\end{tabular}
\caption{Model performance comparison on the MP-Bench dataset. Baseline refers to models fine-tuned using meteorological data from 3 pressure levels. MMLM refers to models fine-tuned using 185 pressure levels and combining the three plug-and-play modules (DTGF, TGS, and TGCA). Scoring scales: First Four Entries (0-100); Last Entry (0-5) . }
\label{tab:combined_results} 
\vspace{-0.4cm}
\end{table*}

\textbf{Regional Severe Weather Questions (RSW)}: These questions are designed to evaluate the model's ability to identify types of severe weather in specific  region. Given a location specified in the query, the model analyzes the corresponding meteorological data to determine the occurrence of severe weather. If no severe weather is detected, “no severe weather" is returned to ensure format consistency and evaluation reliability. 

\textbf{National Severe Weather Questions (NSW)} evaluates the model’s ability to understand and describe nationwide severe weather conditions for a given date. Based on the meteorological data, the model must generate a structured natural language response listing all severe weather events in the form of [$geographic$ $name$][$weather$ $type$][$severity$ $level$]. 

\section{Meteorological Multimodal Large Model}
The proposed MMLM incorporates three plug-and-play modules: Dynamic Time-Gated Fusion (DTGF), Text-Driven Gaussian Spatial Masking (TGS), and Text-Driven Channel Attention (TGCA). As shown in \cref{fig:enter-label1}, the meteorological data are processed in parallel through these modules. Their outputs are concatenated and passed to the fusion layer, where a learnable 3D convolution adaptively integrates temporal, spatial, and channel features into a unified representation, which is then mapped to the baseline model’s input dimensionality through an MLP. The resulting features are fed into an LLM to generate severe weather textual warnings.
\subsection{DTGF}
To better identify spatiotemporal regions with abrupt or substantial changes in the meteorological field, we employ the DTGF module, which dynamically generates gating based on the differences in meteorological data between adjacent time steps and performs weighted fusion of temporal tokens. The module can be formulated as follows:
\begin{gather}
\Delta \mathbf{x}_t = \| \mathbf{x}_t - \mathbf{x}_{t-1} \|_2, \quad t = 2, \dots, T, \\
g_t = \text{Sigmoid} \left( \fn (\Delta \mathbf{x}_t) \right), \quad g_t \in [0, 1], \\
\tilde{\mathbf{x}}_t = g_t \cdot \mathbf{x}_t.
\end{gather}
where $\mathbf{x}_t \in \mathbb{R}^{B \times 1 \times L \times C}$ represents the t-th hour of ERA5 data flattened along the spatial dimension with L = H $\times$ W and C denoting channel dimension. $\Delta \mathbf{x}_t \in \mathbb{R}^{B \times 1 \times L}$ is the L2-norm of adjacent hour data in channel dimension and $\Delta \mathbf{x}_1$ is padded by zeros. The  $g_t \in \mathbb{R}^{B \times 1 \times L \times 1}$ represents gating weight, and is applied to generate the weighted meteorological features $\tilde{\mathbf{x}}_t \in \mathbb{R}^{B \times 1 \times L \times C}$.

\subsection{TGS}
To guide the model to focus on the geographic locations specified in the textual input, we propose TGS module. This module maps the geographic coordinates extracted from text events onto the gridded meteorological data, generates a 2D Gaussian weight mask around each point, and weights the spatial features of each channel and time step to enhance the model’s attention on those regions. The module performs the following operations:

\begin{gather}
(h_i, w_i)
    = \left( \arg\min_{h} |\phi_i - \text{lat}(h)|,\ 
            \arg\min_{w} |\lambda_i - \text{lon}(w)| \right), \\
G_i(h,w)
    = \exp\left( -\frac{(h - h_i)^2 + (w - w_i)^2}{2\sigma^2} \right), \\
M(h, w) 
    = \sum_{i=1}^{N} G_i(h, w).
\end{gather}

Here, $(h_i, w_i)$ are the raster indices mapped from geographic coordinates $(\phi_i, \lambda_i)$ extracted from text events via nearest-grid-point matching, with $h \in [0,H)$, $w \in [0,W)$. In the above equations, $i = 1,2,\dots,N$, where $N$ denotes the number of coordinates from text events, $\sigma$ controls the Gaussian width, $G_i(h,w) \in \mathbb{R}^{H \times W}$ computes the Gaussian weights around location $(h_i, w_i)$, and $M(h,w) \in \mathbb{R}^{H \times W}$ represents the aggregated spatial mask applied to weather features.

\subsection{TGCA}
Since each weather element includes 37 pressure-level channels in the vertical dimension, the aggregation of 185 input channels from the five elements introduces a significant amount of redundancy. To better leverage this high-dimensional information, we propose the TGCA, which dynamically generates attention weights for each weather channel and re-weights the original spatiotemporal features based on the similarity between the input text and channel descriptions. The module can be formulated as follows:

\begin{align}
&\mathbf{P} = \mathrm{Linear}(\mathbf{y}) , \\
&\mathbf{V} = \mathrm{Mean}(\mathbf{X}) ,\\
&\mathbf{Y} = \mathbf{X}\,\cdot\,\text{Sigmoid}\bigl(\operatorname{Softmax}(\mathbf{V}\,\mathbf{P}^\top)\,\mathbf{P}\bigr).
\end{align}

where $\mathbf{y} \in \mathbb{R}^{B \times L' \times D}$ denotes text embeddings of length $L'$ with dimension $D$. $\mathbf{P} \in \mathbb{R}^{B \times L' \times C}$ projects text embeddings to the channel dimension via a learnable linear layer $\linear: \mathbb{R}^D \to \mathbb{R}^C$. $\mathbf{V} \in \mathbb{R}^{B \times C}$ computes channel-wise descriptors by applying spatiotemporal averaging over the input ERA5 data tensor $\mathbf{X} \in \mathbb{R}^{B \times T \times C \times H \times W}$. $\operatorname{Softmax}(\mathbf{V},\mathbf{P}^\top) \in \mathbb{R}^{B \times L'}$ represents attention weights, and sigmoid gating further applies text-guided channel weights to the original meteorological data to obtain refined ERA5 features $\mathbf{Y} \in \mathbb{R}^{B \times T \times C \times H \times W}$, thereby enabling text-conditioned channel feature selection.

\section{Experiments}
\subsection{Experimental Settings }
All experiments were carried out on a distributed cluster of eight NVIDIA A800 GPUs (40GB) for both training and evaluation. We selected four baseline models, namely Qwen2.5-VL-7B-Instruct \cite{bai2025qwen2}, LLaVA-NeXT-Video-7B \cite{zhang2024llavanextvideo}, Video-LLaVA-7B \cite{lin2023video}, and InternVL3-8B \cite{chen2024internvl}, and integrated three plug-and-play modules (DTGF, TGS, and TGCA) for fine-tuning and evaluation. The baseline was fine-tuned with LoRA on all linear layers, and the proposed DTGF, TGCA, and fusion layer were set as learnable components. We used a learning rate of $5\times10^{-5}$, batch size 2, gradient accumulation over 8 steps, and bf16 precision. More details are provided in the appendix.

\subsection{Evaluation Metrics} 
For the T/F and RSW questions, we use Accuracy, which calculates the percentage of samples the model answered correctly. For the MC-main and MC-sub questions, we also calculate Accuracy. However, considering the class imbalance of severe weather events, we also include the Macro-F1 metric to more comprehensively evaluate the model's overall performance across all categories, including rare ones. All Acc and Macro-F1 metrics are scaled to a range of 0-100. For the NSW questions, we adopted an “LLM-as-a-judge" approach \cite{wang2025evaluation,panickssery2024llm,zheng2023judging,liu2023geval}. We first developed a fine-grained scoring criterion in consultation with meteorological experts (See in appendix), and then employed GPT-4o for automated semantic scoring based on this expert-defined rubric, considering completeness, accuracy, and expression quality.

\begin{table*}[!htbp]
\centering
\begin{tabular}{@{}ccc cc cc c c c@{}}
\toprule
\multicolumn{3}{c}{\textbf{MMLM (Qwen2.5\,-\,VL-7B-Instruct)}} &
\multicolumn{2}{c}{\textbf{MC-main}} &
\multicolumn{2}{c}{\textbf{MC-sub}} &
\multicolumn{1}{c}{\textbf{T/F }} &
\multicolumn{1}{c}{\textbf{RSW}} &
\multicolumn{1}{c}{\textbf{NSW }} \\
\cmidrule(lr){1-3}\cmidrule(lr){4-10}
\textbf{DTGF} & \textbf{TGS} & \textbf{TGCA} &
\textbf{Acc $\uparrow$} & \textbf{Macro-F1 $\uparrow$} &
\textbf{Acc $\uparrow$} & \textbf{Macro-F1 $\uparrow$} &
\textbf{Acc $\uparrow$} & \textbf{Acc $\uparrow$} & \textbf{Score $\uparrow$} \\
\midrule
\texttimes & \texttimes & \texttimes & 42.73 & 10.27    & 28.01 & 4.14    & 58.47 & 41.25 & 1.1 \\
$\checkmark$ & \texttimes & \texttimes & 53.28 & 25.36    & 34.84 & 10.36   & 57.23 & 47.30 & 1.2 \\
\texttimes & $\checkmark$ & \texttimes & 44.57 & 22.56    & 29.66 & 9.07    & 54.81 & 41.92 & 1.2 \\
\texttimes & \texttimes & $\checkmark$ & 50.91 & 22.48    & 34.62 & 9.20    & 60.35 & 50.29 & 1.4 \\
$\checkmark$ & $\checkmark$ & \texttimes & 53.81 & 23.50   & 32.02 & 9.64   & 56.63 & 54.02 & 1.4 \\
$\checkmark$ & \texttimes & $\checkmark$ & 50.93 &  24.93  & 32.61 & 9.99    & 67.17 & 51.87 & 1.3 \\
\texttimes & $\checkmark$ & $\checkmark$ & 51.97 & 24.98 & 31.99 & 10.42 & 70.44 & 57.23 & 1.4 \\
$\checkmark$ & $\checkmark$ & $\checkmark$ & \textbf{58.27} & \textbf{26.19}    & \textbf{47.37} & \textbf{11.83}    & \textbf{79.21} & \textbf{58.63} & \textbf{1.7} \\
\bottomrule
\end{tabular}
\vspace{-0.1cm}
\caption{Ablation study of the three proposed plug-and-play modules in MMLM.}
\label{tab:model_comparisons}
\vspace{-0.5cm}
\end{table*}

\subsection{Quantitative Results Analysis}
\subsubsection{Closed-source Model Performance Evaluation}
We selected GPT-4o as a representative closed-source model and systematically evaluated its performance in severe weather forecasting to establish the critical necessity of the proposed MMLM. Inspired by the CLLMate \cite{li2024cllmate}, this study selected four key meteorological variables, including 2-meter temperature, 10-meter u-component of wind, 10-meter v-component of wind, and total precipitation—and vectorially synthesized the u/v wind components into a unified wind field to construct a three-channel meteorological input compatible with GPT-4o's requirements. Concurrently, ERA5 reanalysis data from 12 hours post-severe weather warning issuance were spatiotemporally matched with corresponding warning texts. 
As shown in \cref{tab:combined_results}, The results were poor across all metrics, especially on the T/F task, where the Acc was only 0.19\%, a score far below the level of random guessing. Furthermore, its Macro-F1 scores on the MC-main and MC-sub tasks were merely 2.92\% and 0.88\%, respectively. This indicates the model has almost no capability to distinguish rare categories. This confirms that closed-source models without fine-tuning lack the capability to understand meteorological data.

\subsubsection{Performance Comparison of Baseline Models}
To directly evaluate the open-source models' understanding of meteorological data, we conducted a comparative experiment. In this experiment, we adopted the same 3-channel meteorological data as the closed-source baseline (mentioned in Section 5.3.1) and used it to fine-tune the four open-source models. As shown in \cref{tab:combined_results}, the performance of the Baseline category, which uses 3-channel data to fine-tune open-source models, shows a fundamental improvement compared to the closed-source model. For example, LLaVA-NeXT-Video-7B achieved a T/F Acc of 69.23\%, while GPT-4o scored only 0.19\%.

However, baseline models showed low Macro-F1 scores. This limitation can be attributed to the inherent class imbalance in the dataset and the reliance of certain weather type, such as hail, on higher-resolution meteorological inputs that open-source models are difficult to capture using the 3-channel  meteorological inputs. These findings highlight the need for more adaptive fusion mechanisms to enhance feature discrimination among diverse weather types.

\subsubsection{Performance Comparison of MLLMs}
To further evaluate the effectiveness of the proposed MMLMs, we conducted a comparison between the MMLM-enhanced models and their baseline counterparts. As shown in \cref{tab:combined_results}, all MMLM variants consistently outperformed the baselines across every task, demonstrating the effectiveness of the multimodal fusion strategy.

Among them, the Qwen2.5-VL-based MMLM achieved the best overall performance. Compared with its baseline (MC-main accuracy 56.26\%, Macro-F1 26.88\%), this represents a substantial improvement of over 16 percentage points in accuracy and nearly double the Macro-F1 score. Notably, the NSW task also shows a clear improvement over the baseline, but its relatively low absolute score indicates that this task remains highly challenging and still offers considerable room for further improvement. This performance gain can be attributed to the three proposed plug and play modules, namely DTGF, TGS, and TGCA, which enable dynamic fusion across temporal, spatial, and vertical dimensions. Furthermore, the Qwen2.5-VL architecture’s strong pretrained backbone and efficient 2D RoPE with windowed attention mechanism ensure better alignment between meteorological features and textual queries, leading to more accurate reasoning and robust severe weather event prediction. 

\begin{figure*}[!htbp]
    \centering
    \includegraphics[width=1.0\textwidth]{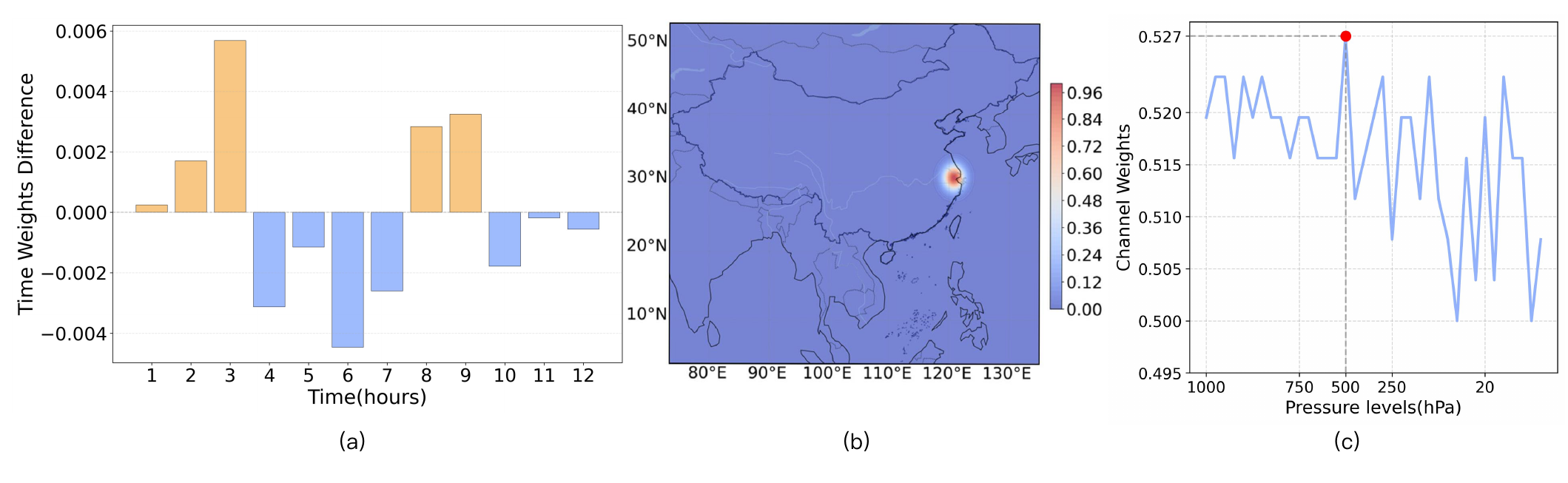}
    \vspace{-0.89cm}
    \caption{Weight distribution patterns of three types of plug-and-play modules: (a) DTGF temporal weights difference (positive: higher for red warnings; negative: higher for blue warnings); (b) TGS spatial attention map;(c) TGCA channel weights of the V-component of wind across pressure levels. Additional examples are in appendix.}
    \label{fig:xiaorong}
    \vspace{-0.2cm}
\end{figure*}

\subsection{Ablation Study}
\subsubsection{Ablation Study on Plug-and-Play Modules}
We conducted ablation studies on the three plug and play modules (DTGF, TGS, and TGCA) using 5,000 samples from the MP-Bench dataset to ensure consistency with the training data. The following analysis examines the individual contribution and combined effects of each module within the MMLM framework.

\textbf{DTGF significantly enhances model's temporal feature extraction ability.}
With the DTGF module operating in standalone mode, the model achieved significant improvement in MC questions aimed at identifying severe weather types and severity levels. \cref{fig:xiaorong} (a) displayed the temporal weight difference for blue and red warnings. For red rainstorm warnings, the module assigned higher weights to the first three hours after issuance than to other periods, consistent with the physical evolution of meteorological fields (verified in the appendix). This indicated that the module could effectively focus on meteorological information during critical time windows, thereby improving the accuracy of severe weather event forecasts.

\textbf{TGS improves the model's focus on the geographic area specified by the text.}
\cref{fig:xiaorong} (b) visually demonstrated the Gaussian weight distribution of the TGS module, revealing that the model precisely focused on regions specified in queries. When TGS and DTGF were combined, they significantly improved performance on both MC-main and RSW, confirming the complementary nature of their features.

\textbf{TGCA filters redundant channels to allow models to understand the meteorological factors that influence severe weather.}
When the TGCA module was introduced, T/F metrics showed significant improvement, with the TGCA-TGS combination yielding the highest gain (11.97\%). Additionally, we visualized the 37-channel weight distributions for five meteorological variables (temperature, humidity, precipitation, wind speed, and pressure). Using radial wind speed in \cref{fig:xiaorong} (c) as an example, the wind speed weight at 500 hPa was significantly higher than at other layers, indicating its key role in severe weather event prediction. Channel weight distributions for the remaining four variables were detailed in the appendix. 

\textbf{The parallel fusion of three modules achieves optimal overall performance.}
DTGF’s temporal highlighting effectively complemented TGCA’s adaptive channel reweighting, while TGS spatial masks further guided the model to dynamically concentrate computations on meteorologically high-risk regions. The resulting 19.36\% overall gain in MC-sub strongly confirmed the modules’ complementary functional roles and clearly demonstrated the robustness and effectiveness of the proposed parallel module fusion strategy.

\subsubsection{Analysis of Temporal Window Length}
To investigate the optimal temporal window for aligning severe weather events with ERA5 data, we conducted an ablation study. We compared the 12-hour window [$t$, $t$+11] used by our main model against shorter 1h, 3h, and 6h windows. These models were trained on the 5000-sample subset and evaluated on the MC dataset, with additional RSW and NSW results reported in the appendix. As shown in \cref{tab:ablation_window} for MC and in the appendix for RSW and NSW, the results demonstrate that the 12-hour window significantly outperforms the shorter windows, indicating that the proposed DTGF module can effectively capture useful temporal information from longer time spans. Considering that 12 hours is sufficient to capture the key dynamic processes of most severe weather events, and that longer windows would introduce prohibitive computational overhead, we consider the 12-hour window a near-optimal balance between model performance and computational efficiency.
\begin{table}[h]
\centering
\resizebox{\columnwidth}{!}{
\begin{tabular}{@{}ccccc@{}} 
\toprule
\textbf{Window Length} & \multicolumn{2}{c}{\textbf{MC-main}} & \multicolumn{2}{c}{\textbf{MC-sub}} \\
\cmidrule(r){2-5} 
 & \textbf{Acc ↑} & \textbf{Macro-F1 ↑} & \textbf{Acc ↑} & \textbf{Macro-F1 ↑} \\
\midrule
{[ $t$ , $t$+1 )} & 54.29 & 21.19 & 42.77 & 8.83 \\
{[ $t$ , $t$+5 ]} & 58.01 & 24.00 & 46.83 & 10.00 \\
{[ $t$ , $t$+11 ]} & \textbf{58.27} & \textbf{26.19} & \textbf{47.37} & \textbf{11.83} \\
\bottomrule
\end{tabular}
}
\caption{Analysis of Temporal Window Lengths (1h, 6h, 12h).}
\label{tab:ablation_window}
\vspace{-0.45cm}
\end{table}

\subsubsection{Cross-geographical Generalization}
We employed the best-performing Qwen2.5-VL–based MMLM to evaluate cross-regional generalization. The model was trained on the China-based MP-Bench dataset and directly tested on 1,000 samples from the U.S. NOAA Storm Events Database to assess external validity across differing geographical and climatic domains. As the NOAA dataset contains only categorical weather types, the cross-regional evaluation was formulated as an MC-main task covering the major categories. Under the same evaluation setting, the proposed MMLM notably outperformed the baseline models. Although performance declined due to regional and climatic differences, the model still achieved 64.20\% accuracy and 35.18\% Macro-F1, demonstrating effective cross-regional generalization.

\section{Conclusion}
AI-driven severe weather event prediction has shown great potential. In this study, we have collected the MP-Bench dataset, thereby addressing the scarcity of data for severe weather forecasting and introducing a more precise alignment strategy between meteorological grids and textual warnings. Building on this foundation, we have developed a Meteorological Multimodal Large Model (MMLM) and have integrated three plug-and-play modules into its architecture, markedly enhancing the model’s ability to capture spatiotemporal patterns and vertical pressure‑level information.  Nevertheless, a case study of misclassified samples shows that they are associated with more complex meteorological field patterns, as detailed in the appendix. This suggests that integrating physical constraints with multi-source data will be important for further improving forecasting accuracy in such complex scenarios.

{
    \small
    \bibliographystyle{ieeenat_fullname}
    \bibliography{main}
}

\newpage
\appendix
\begin{center}
{\Large\bfseries Supplementary Material\par}
\end{center}
\setcounter{section}{0}
\renewcommand\thesection{\arabic{section}} 
\section{Dataset Details}
\subsection{Meteorological data}
In this study, to investigate the impact of atmospheric physical processes in the near-surface layer, troposphere, and stratosphere on warning issuance, we selected five key variables from the ERA5 reanalysis dataset across 37 vertical pressure levels in \cref{tab:era5}. Specifically, the 800–1000 hPa range represents the near-surface layer, 200–800 hPa corresponds to the main troposphere, and levels below 200 hPa are associated with the stratosphere.

\subsection{Severe Weather Event Distribution}
\begin{figure}[htbp]
    \begin{subfigure}[b]{\columnwidth}
        \centering
        \includegraphics[width=1.0\textwidth]{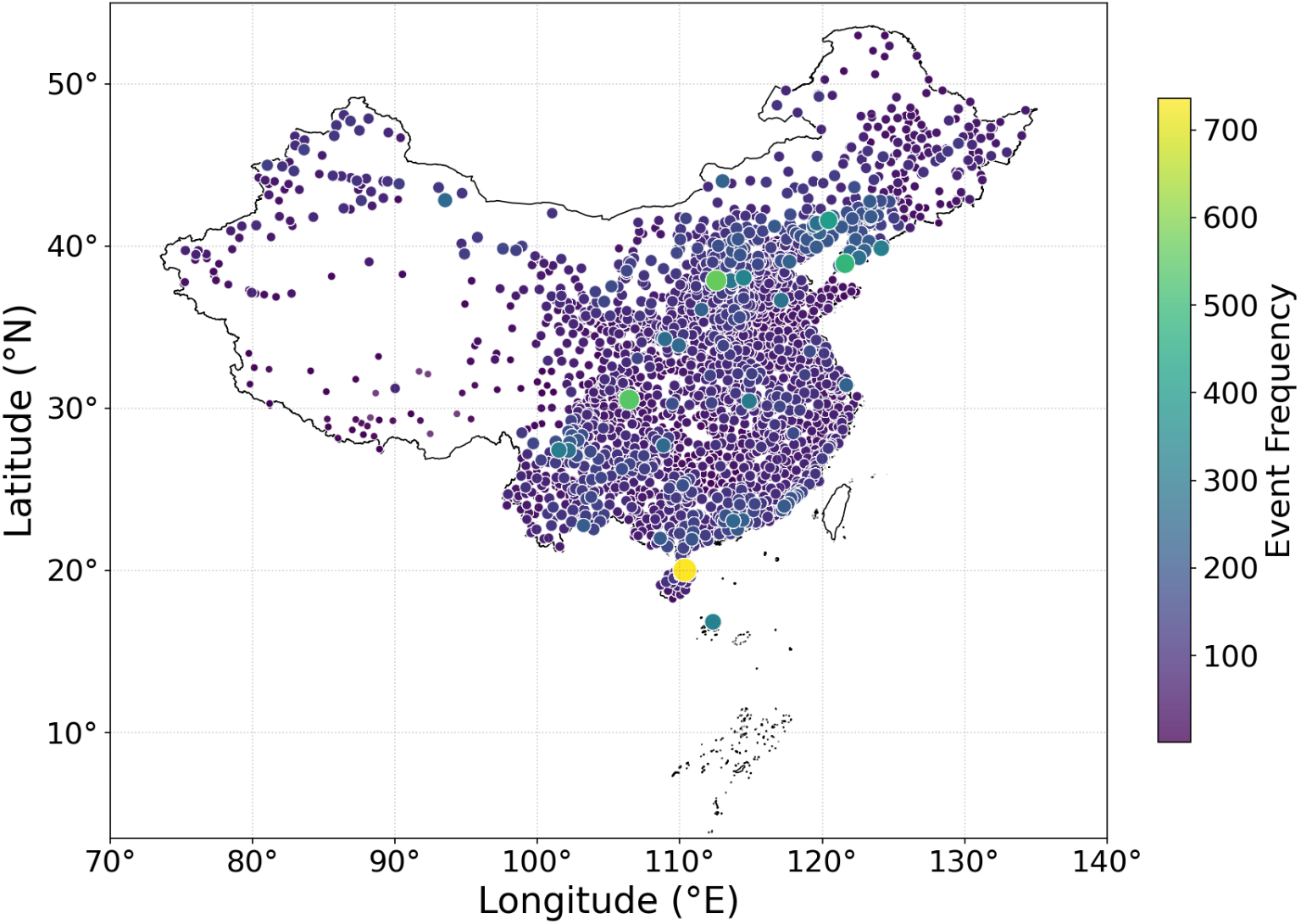} 
        \caption{Spatial distribution of severe weather events across China. Symbol size corresponds to regional severe weather frequency.} 
        \label{subfig:1} 
    \end{subfigure}
    \begin{subfigure}[b]{\columnwidth}
        \centering
        \includegraphics[width=1.0\textwidth]{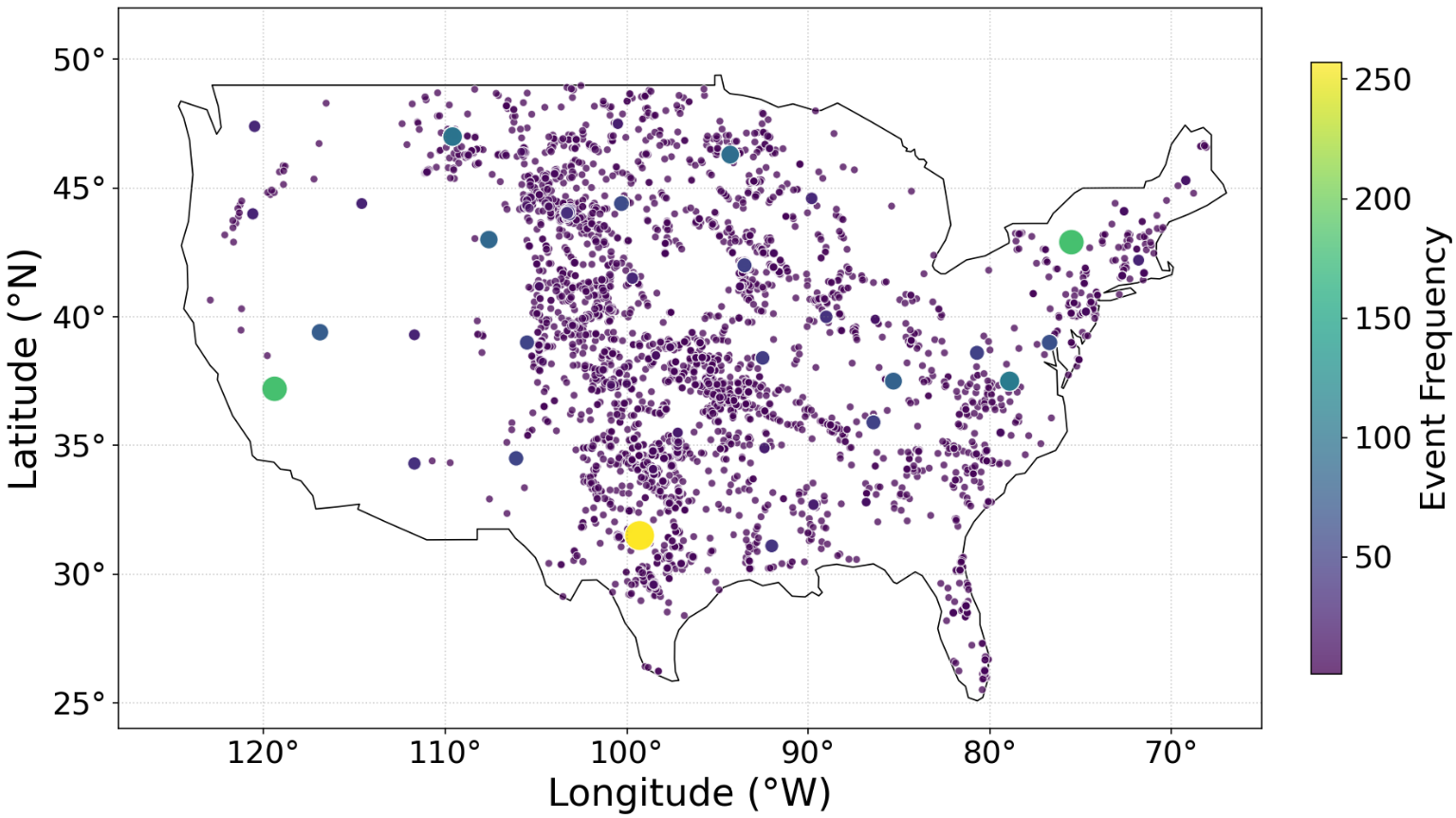}
        \caption{Spatial distribution of severe weather events across US. Symbol size corresponds to regional severe weather frequency.}
        \label{subfig:2}
    \end{subfigure}
    \caption{Spatial distribution of severe weather events of China and US.} 
    \label{fig:region}
\end{figure}

\cref{fig:region} (a) displays the severe weather event data for the China region, which serves as the training and testing set for the model. The dots represent the locations of the events, with color and symbol size corresponding to the regional event frequency, clearly indicating that the densely populated eastern and southeastern regions of China are areas with high frequencies of severe weather.

\cref{fig:region} (b) depicts the distribution of severe weather events in the US region. Specifically, this US subset consists of 1,000 samples drawn from the NOAA Storm Events database and covers all seasons, with the proportions of each severe-weather type matched to those of the MP-Bench test set (see \cref{tab:text_event}). This data is designated as the generalization validation set to test the model's ability to generalize to different geographical areas. Similar to the China data, the plot uses color and size to show event frequency, highlighting the Midwestern and Eastern parts of the US as regions prone to high severe weather occurrences.

\begin{figure}[htbp]
    \begin{subfigure}[b]{\columnwidth}
        \centering
        \includegraphics[width=1.0\textwidth]{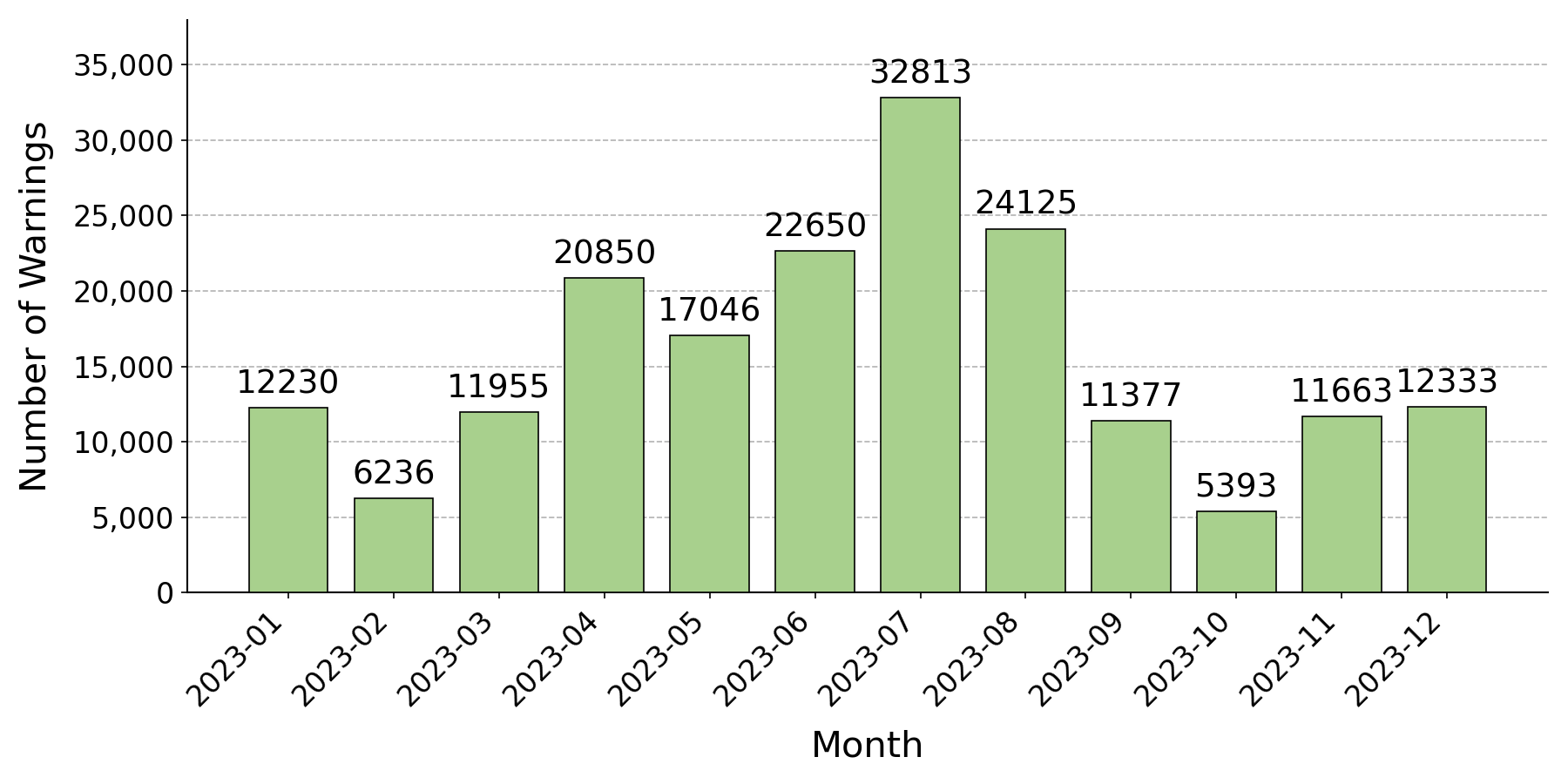} 
        \caption{2023 severe weather events bar chart.} 
        \label{subfig:1} 
    \end{subfigure}
    \begin{subfigure}[b]{\columnwidth}
        \centering
        \includegraphics[width=1.0\textwidth]{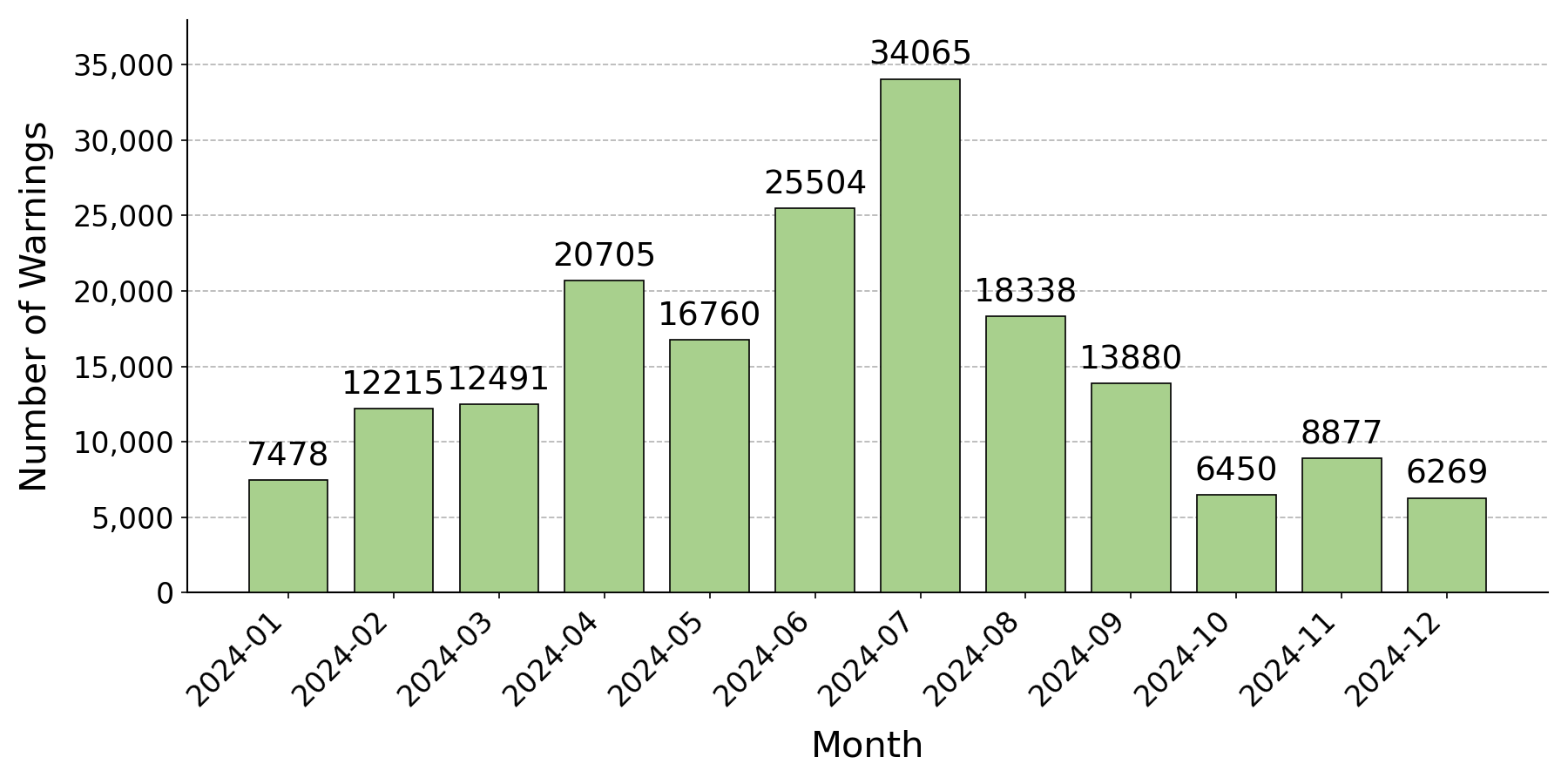}
        \caption{2024 severe weather events bar chart.}
        \label{subfig:2}
    \end{subfigure}
    \caption{Monthly bar charts of severe weather events for the years (a) 2023 and (b) 2024.} 
    \label{fig:total}
\end{figure}

\subsection{Train-Test Split Description}
To construct a temporally independent evaluation protocol, we adopt a year-wise dataset split, using 2023 as the training set and 2024 as the test set. \cref{tab:text_event} summarizes the distribution of severe weather warnings for both years. Although the absolute event counts vary across years, the relative proportions of different severe weather types remain highly consistent, providing a stable data foundation for training–testing separation. Gale accounts for the largest proportion in both years (2023: 43.54\%, 2024: 41.52\%), ensuring sufficient samples for model learning and robust evaluation. Rain Storm and Heat Wave show similar ratios across years (22.24\% vs. 23.50\%; 9.74\% vs. 11.91\%), enabling reasonable generalization on mid-frequency categories. Cold Wave, Hail, Frost, and Snow Storm remain low-frequency but stable (each $<5\%$), which is important for evaluating the model’s ability to detect rare severe weather events. The number of Normal samples is almost identical between the two years (about 25k), providing a consistent baseline for distinguishing severe vs. normal weather. 

\subsection{QA types}
MP-Bench comprises four types of QA pairs, with \cref{fig:tf_question}–\cref{fig:nsw_question} illustrating representative examples of each type.

\begin{table}[!htbp]
    \centering
    \begin{tabular}{llrr}
    \toprule
    Year & Type & Num & Ratio (\%) \\
    \midrule
    \multirow{9}{*}{2023} 
        & Gale       &  92939 & 43.54 \\
        & Rain Storm &  47470 & 22.24 \\
        & Heat Wave  &  20787 &  9.74 \\
        & Cold Wave  &  11306 &  5.30 \\
        & Frost      &   7171 &  3.36 \\
        & Hail       &   5794 &  2.71 \\
        & Snow Storm &   3204 &  1.50 \\
        & Normal     &  24810 & 11.62 \\
        & \textbf{All} & \textbf{213481} & \textbf{100.0} \\
\midrule
    \multirow{9}{*}{2024} 
        & Gale       &  86314 & 41.52 \\
        & Rain Storm &  48870 & 23.50 \\
        & Heat Wave  &  24767 & 11.91 \\
        & Cold Wave  &   8200 &  3.94 \\
        & Hail       &   7004 &  3.37 \\
        & Frost      &   4410 &  2.12 \\
        & Snow Storm &   3467 &  1.67 \\
        & Normal     &  24850 & 11.95 \\
        & \textbf{All} & \textbf{207882} & \textbf{100.0} \\
    \bottomrule
    \end{tabular}
    \caption{Statistics of severe weather warnings (2023--2024).}
    \label{tab:text_event}
\end{table}

\begin{table*}[!htbp]
    \centering
    
    \begin{tabular}{llll}
        \toprule
        Variable & Definition & Unit & Pressure Levels (hPa) \\
        \midrule
        z    & geopotential                               & gpm            &         1, 2, 3, 5, 7, 10, 20, 30, 50, 70, 100,\\
        u    & U-component Wind Speed                     & $\mathrm{m/s}$ & 
                125, 150, 175, 200, 225, 250, 300, 350,\\
        v    & V-component Wind Speed                     & $\mathrm{m/s}$ & 
                400, 450, 500, 550, 600, 650, 700, 750,\\
        t    & Temperature                                & K              &         775, 800, 825, 850, 875, 900, 925, 950,\\
        q    & Specific Humidity                          & $\mathrm{kg/kg}$ & 975, 1000   \\
        \bottomrule
    \end{tabular}
    \captionof{table}{Summary of the 5 physical variables in the dataset.}
    \label{tab:era5}
\end{table*}

\section{Experiment Settings}
In MMLM framework, we adopt four base models, Qwen2.5-VL-7B-Instruct, LLaVA-NeXT-Video-7B, Video-LLaVA-7B, and InternVL3-8B, and attach three plug-and-play modules (DTGF, TGS, and TGCA) for fine-tuning and evaluation. All linear layers in the base models are fine-tuned with LoRA, while DTGF, TGCA, and the fusion layer are trained as additional learnable components. Unless otherwise specified, all reported results are averaged over three independent runs. The detailed training and testing 
settings are summarized in \cref{tab:training_params} and \cref{tab:evaluation_params}.

\section{Supplementary Experiment}
\subsection{Analysis of Temporal Window Length}
Building on the temporal-window ablation described in the main text, we further report detailed results for the T/F, RSW and NSW tasks in the appendix. Using the same 5{,}000 samples subset and the three temporal windows [$t$, $t$+1), [$t$, $t$+5], and [$t$, $t$+11] as in the MC experiment, we train separate models and evaluate their performance on T/F, RSW and NSW. As shown in \cref{ablation_window}, the 12-hour window [$t$, $t$+11] consistently achieves the best accuracy across both tasks, confirming that the longer temporal context is beneficial not only for MC but also for regional selection and open-ended description of severe weather events.
\begin{table}[h]
\centering
\resizebox{\columnwidth}{!}{
\begin{tabular}{@{}cccc@{}} 
\toprule
\textbf{Window Length} & \textbf{T/F Acc ↑} & \textbf{RSW Acc ↑} & \textbf{NSW Score ↑} \\
\midrule
{[ $t$ , $t$+1 )}  & 67.17 & 52.71 & 1.1 \\
{[ $t$ , $t$+5 ]}  & 72.23 & 54.63 & 1.6 \\
{[ $t$ , $t$+11 ]} & \textbf{79.21} & \textbf{58.63} & \textbf{1.7} \\
\bottomrule
\end{tabular}
}
\caption{Analysis of Temporal Window Lengths (1h, 6h, 12h).}
\label{ablation_window}
\vspace{-0.45cm}
\end{table}

\section{Physically Consistent in the TGCA Module}
\begin{figure}[htbp]
    \begin{subfigure}[b]{0.45\columnwidth} 
        \centering
        \includegraphics[width=1.0\textwidth]{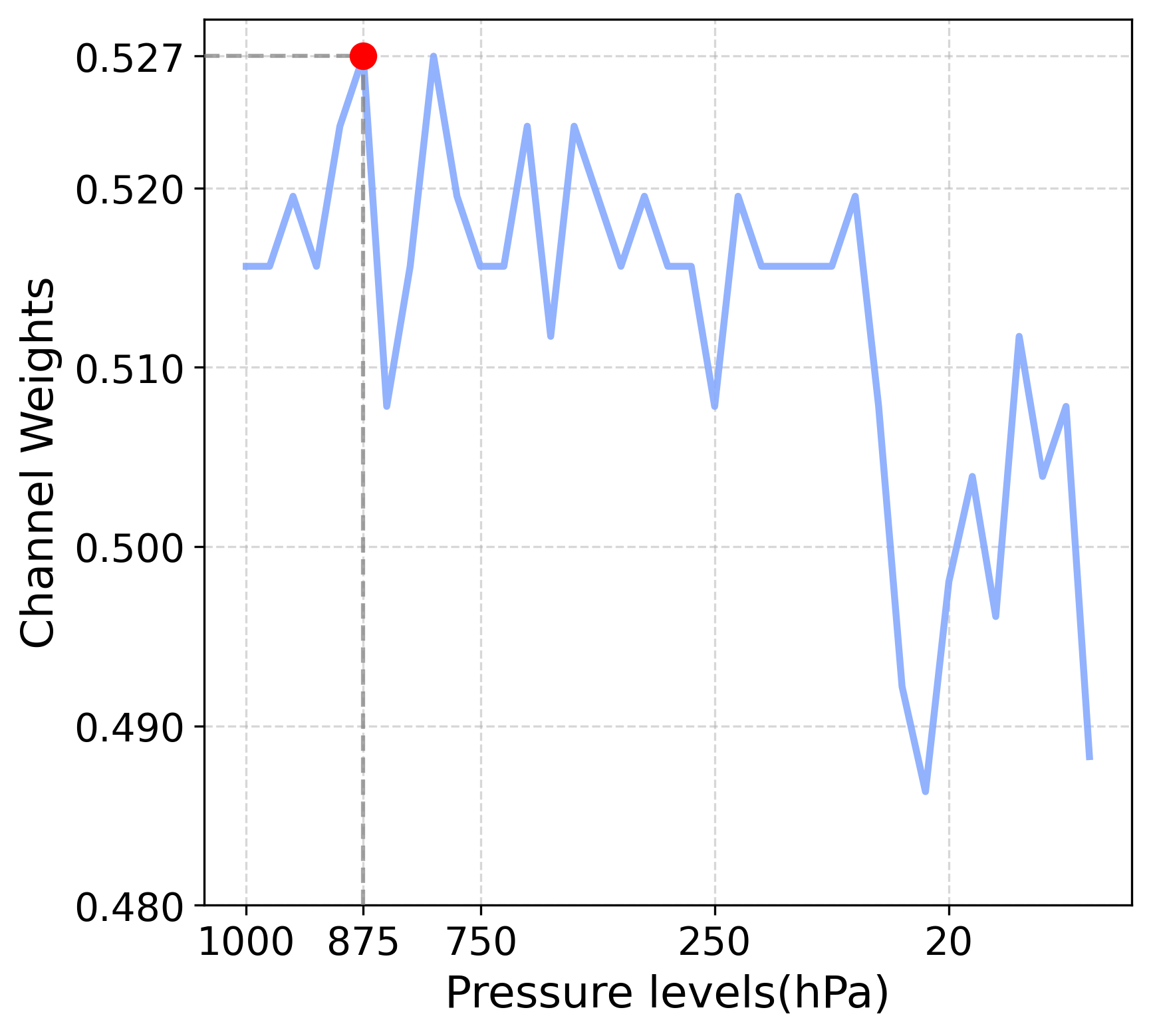}
        \caption{Specific humidity.}
        \label{subfig:2x2a}
    \end{subfigure}
    \begin{subfigure}[b]{0.45\columnwidth}
        \centering
        \includegraphics[width=1.0\textwidth]{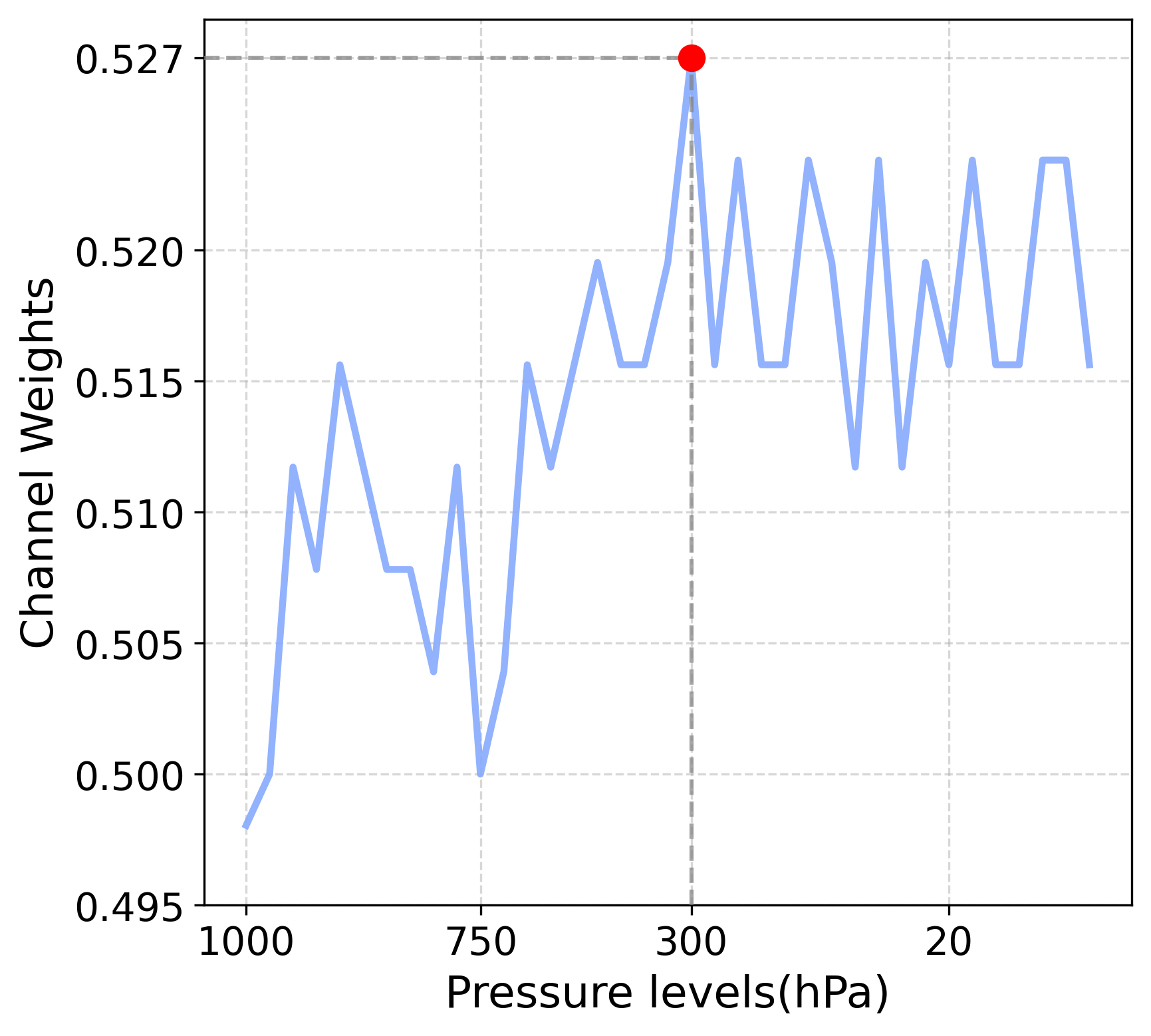}
        \caption{Temperature.}
        \label{subfig:2x2b}
    \end{subfigure}

    \begin{subfigure}[b]{0.45\columnwidth}
        \centering
        \includegraphics[width=1.0\textwidth]{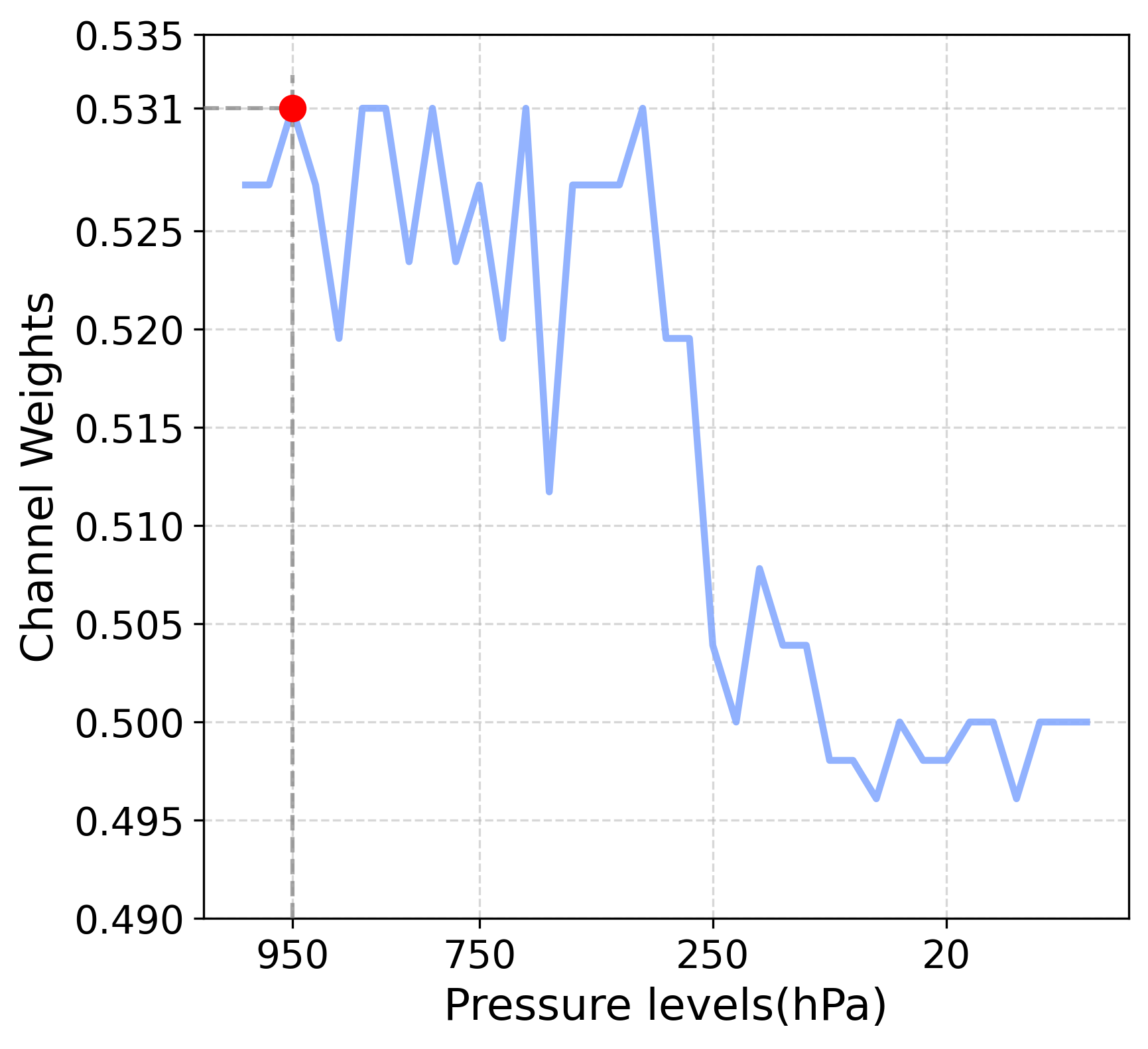}
        \caption{U-component of wind.}
        \label{subfig:2x2c}
    \end{subfigure}
    \begin{subfigure}[b]{0.45\columnwidth}
        \centering
        \includegraphics[width=1.0\textwidth]{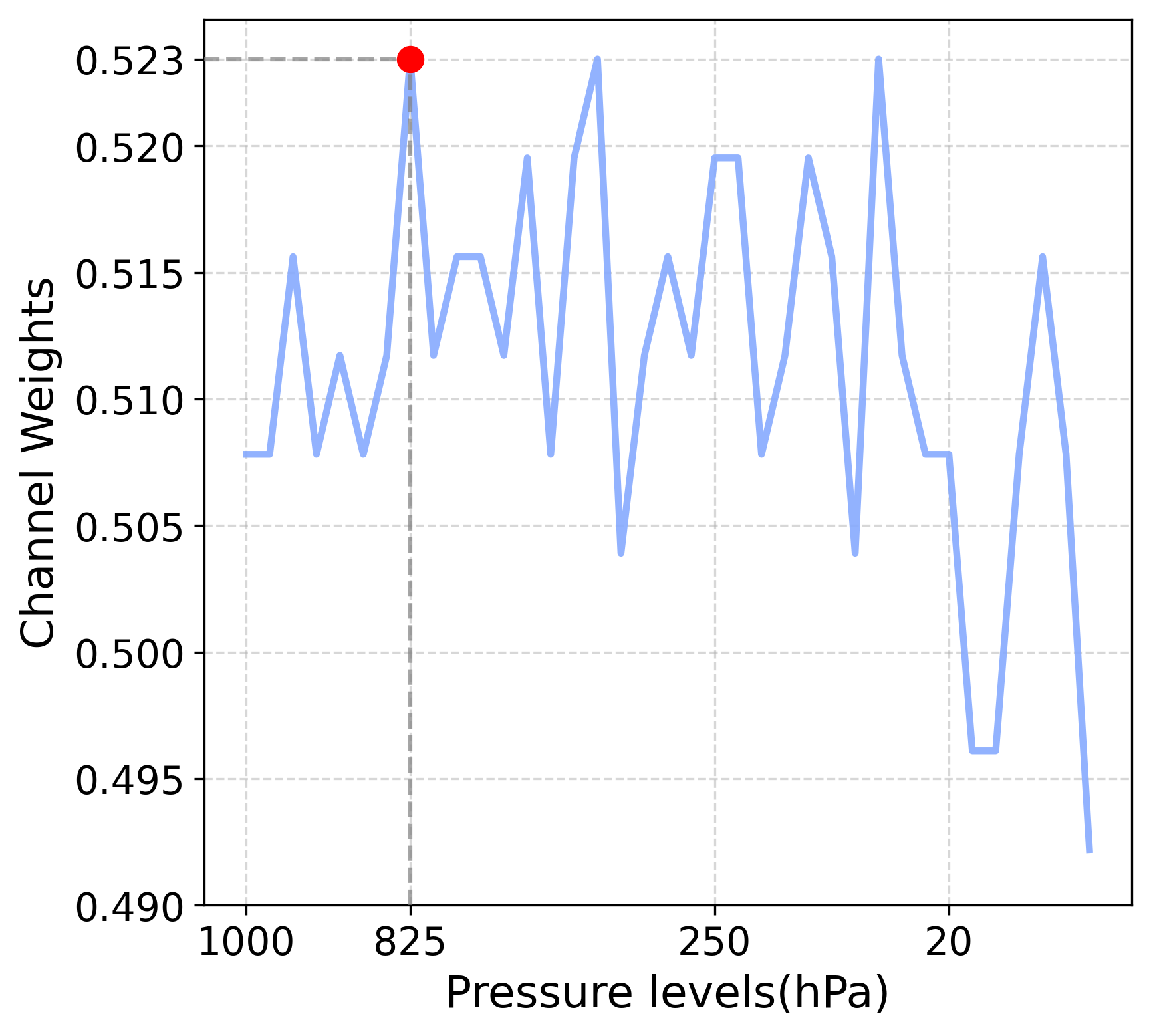}
        \caption{Geopotential.}
        \label{subfig:2x2d}
    \end{subfigure}
    
    \caption{Examples of TGCA's Weight distribution patterns. Each subplot represents a type of severe weather event.}
    \label{fig:2x2}
\end{figure}

To determine the most sensitive pressure levels for five meteorological variables in the ERA5 dataset, we selected seven typical severe weather types (10 samples each) and analyzed them by calculating the average weights across all pressure levels. As shown in \cref{fig:2x2}, the channel-wise attention learned by the TGCA module automatically concentrates on physically meaningful pressure levels: 500hPa meridional wind for mid-tropospheric trough–ridge patterns, 950 hPa zonal wind for near-surface gale-related flows, 300 hPa temperature for upper-level cold cores and jet structures, and geopotential height and humidity around 825–875 hPa for low-level baroclinicity and moisture supply. This alignment with classical synoptic-scale analysis suggests that the model has discovered physically consistent diagnostics.
\section{Physically Consistent in the DTGF Module}
\begin{figure}[htbp!]
    \begin{subfigure}[b]{0.45\columnwidth} 
        \centering
        \includegraphics[width=1.0\textwidth]{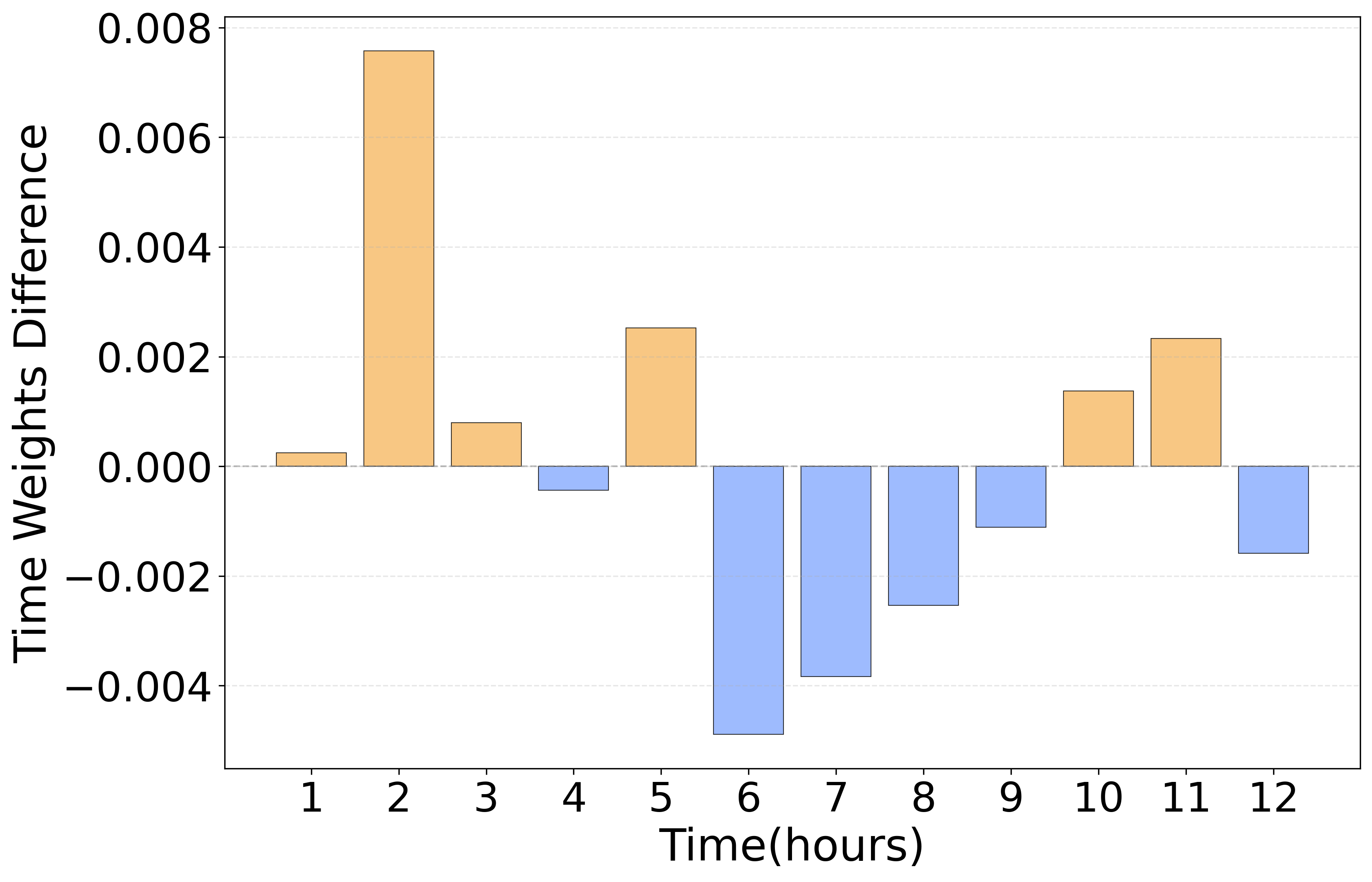}
        \caption{Heat wave.}
        \label{subfig:2x2a}
    \end{subfigure}
    \begin{subfigure}[b]{0.45\columnwidth}
        \centering
        \includegraphics[width=1.0\textwidth]{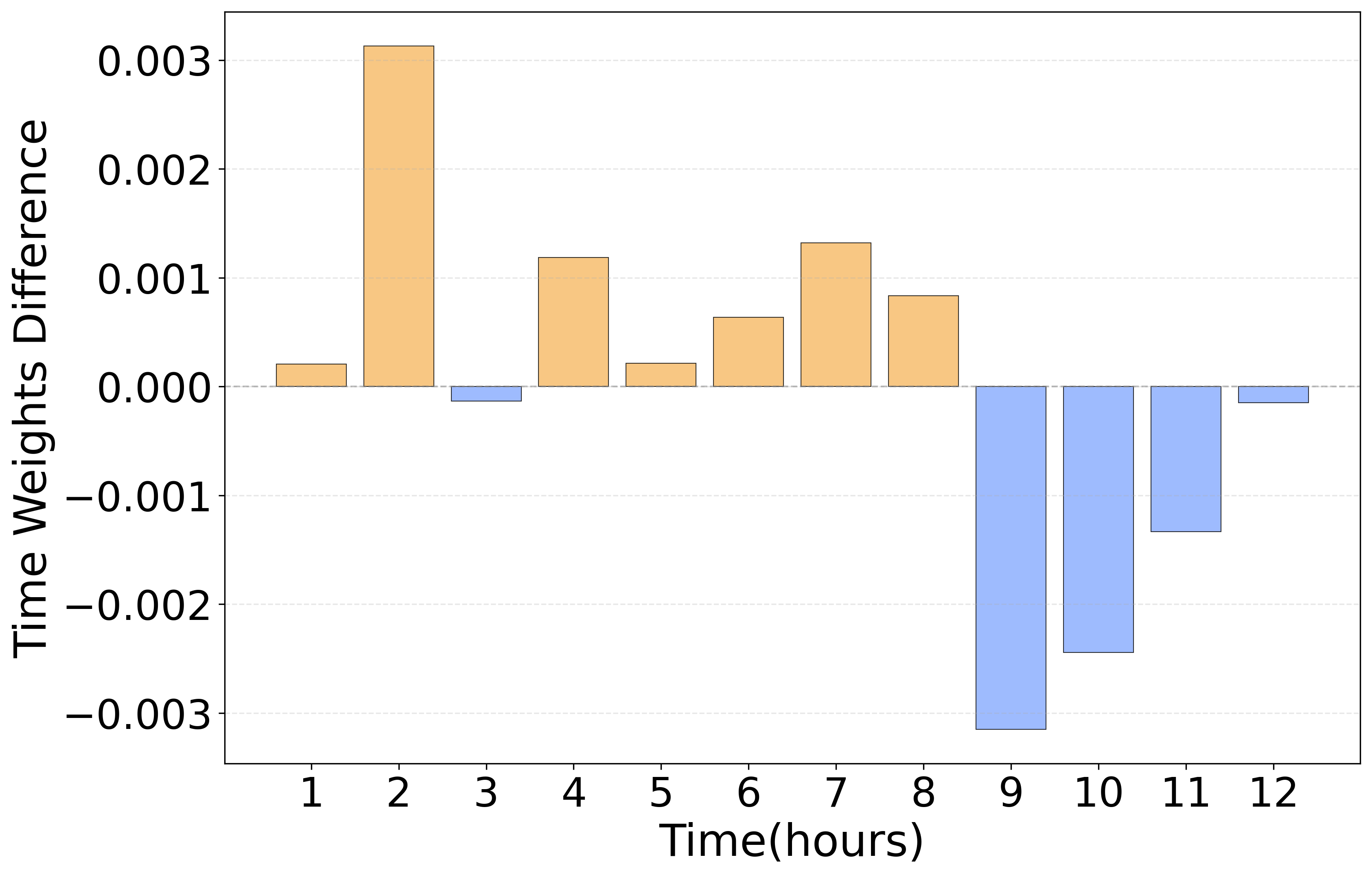}
        \caption{Snow storm.}
        \label{subfig:2x2b}
    \end{subfigure}

    \begin{subfigure}[b]{0.45\columnwidth}
        \centering
        \includegraphics[width=1.0\textwidth]{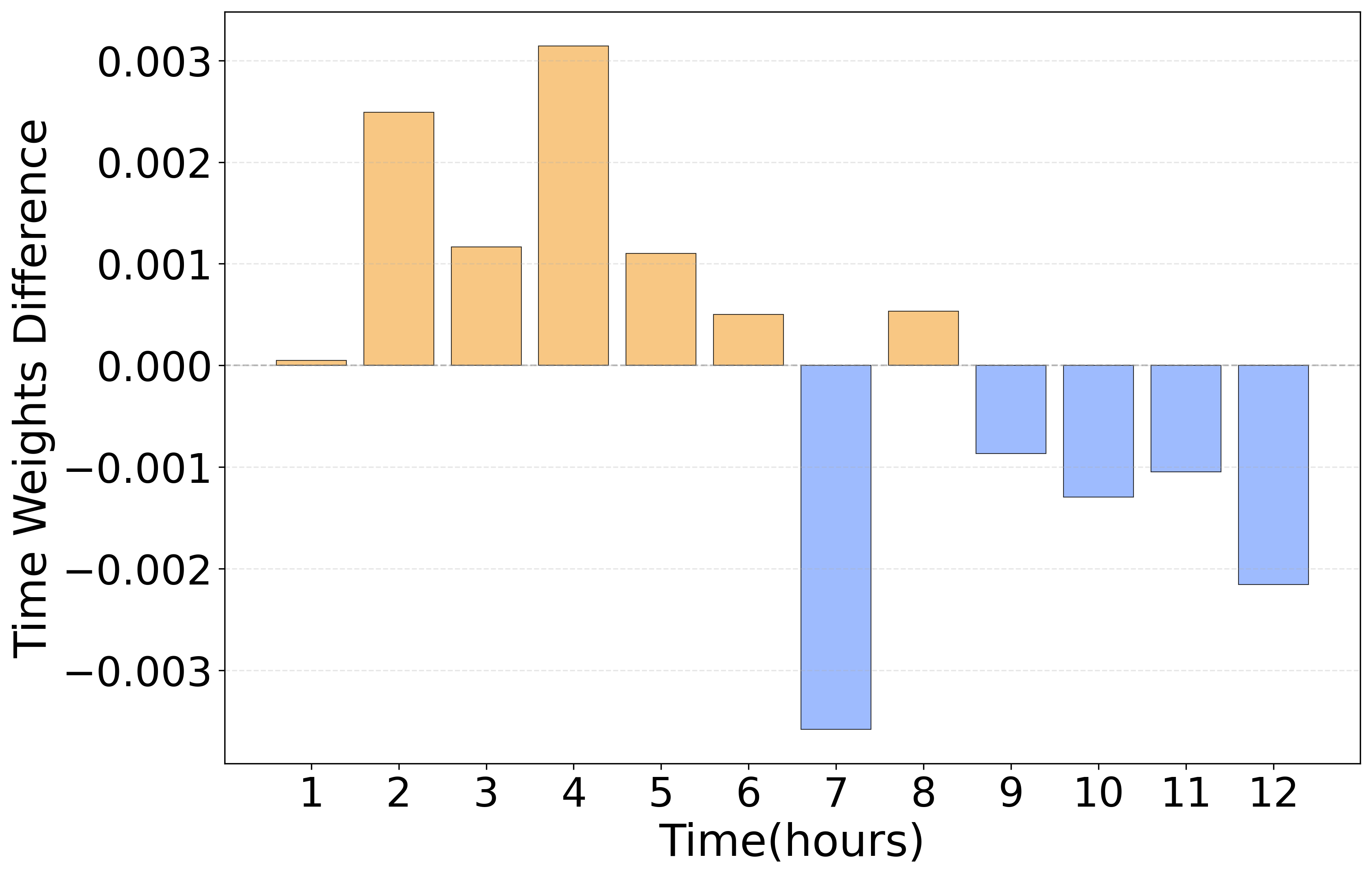}
        \caption{Gale.}
        \label{subfig:2x2c}
    \end{subfigure}
    \begin{subfigure}[b]{0.45\columnwidth}
        \centering
        \includegraphics[width=1.0\textwidth]{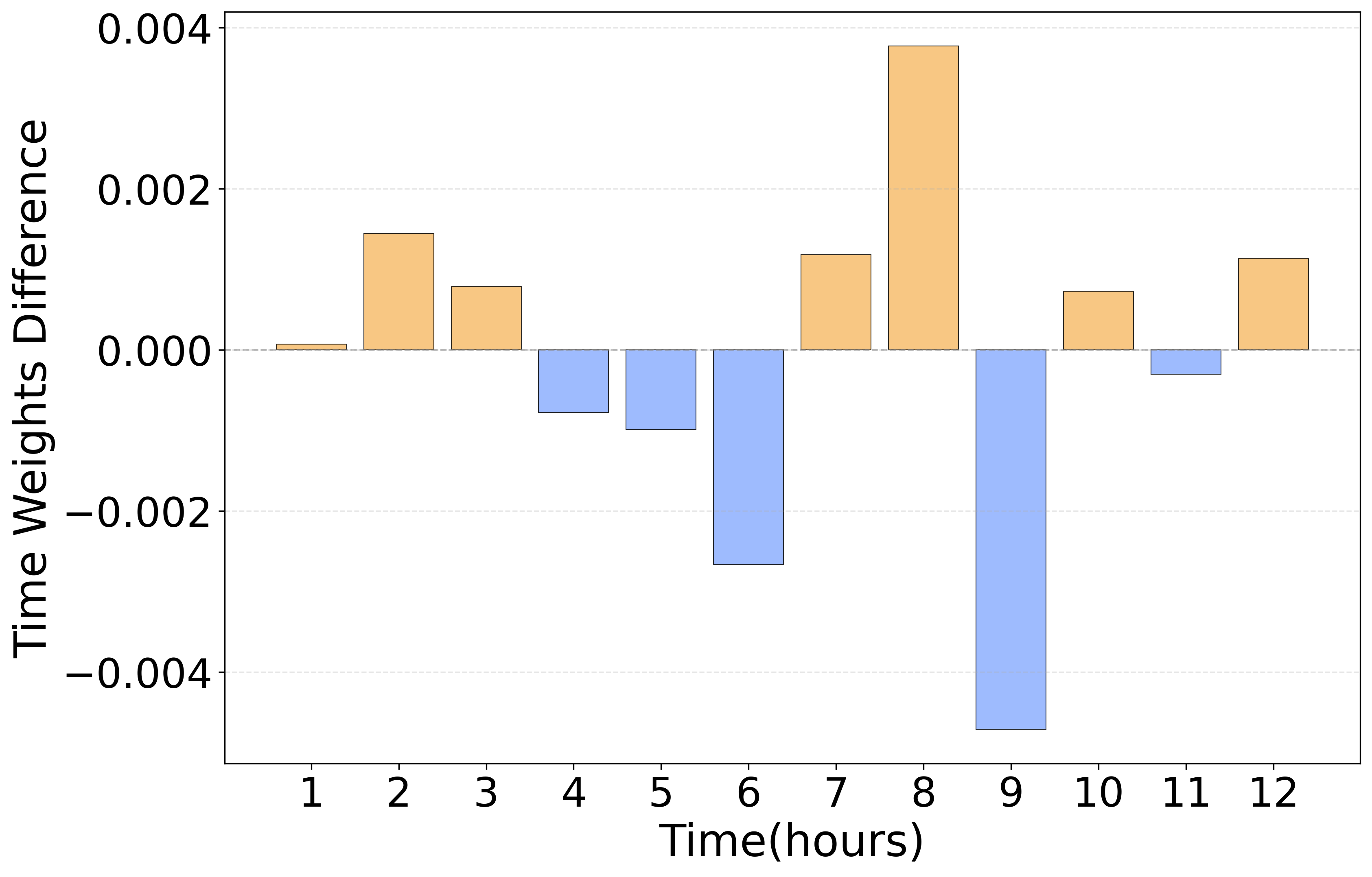}
        \caption{Cold Wave.}
        \label{subfig:2x2d}
    \end{subfigure}
    
    \begin{subfigure}[b]{0.45\columnwidth}
        \centering
        \includegraphics[width=1.0\textwidth]{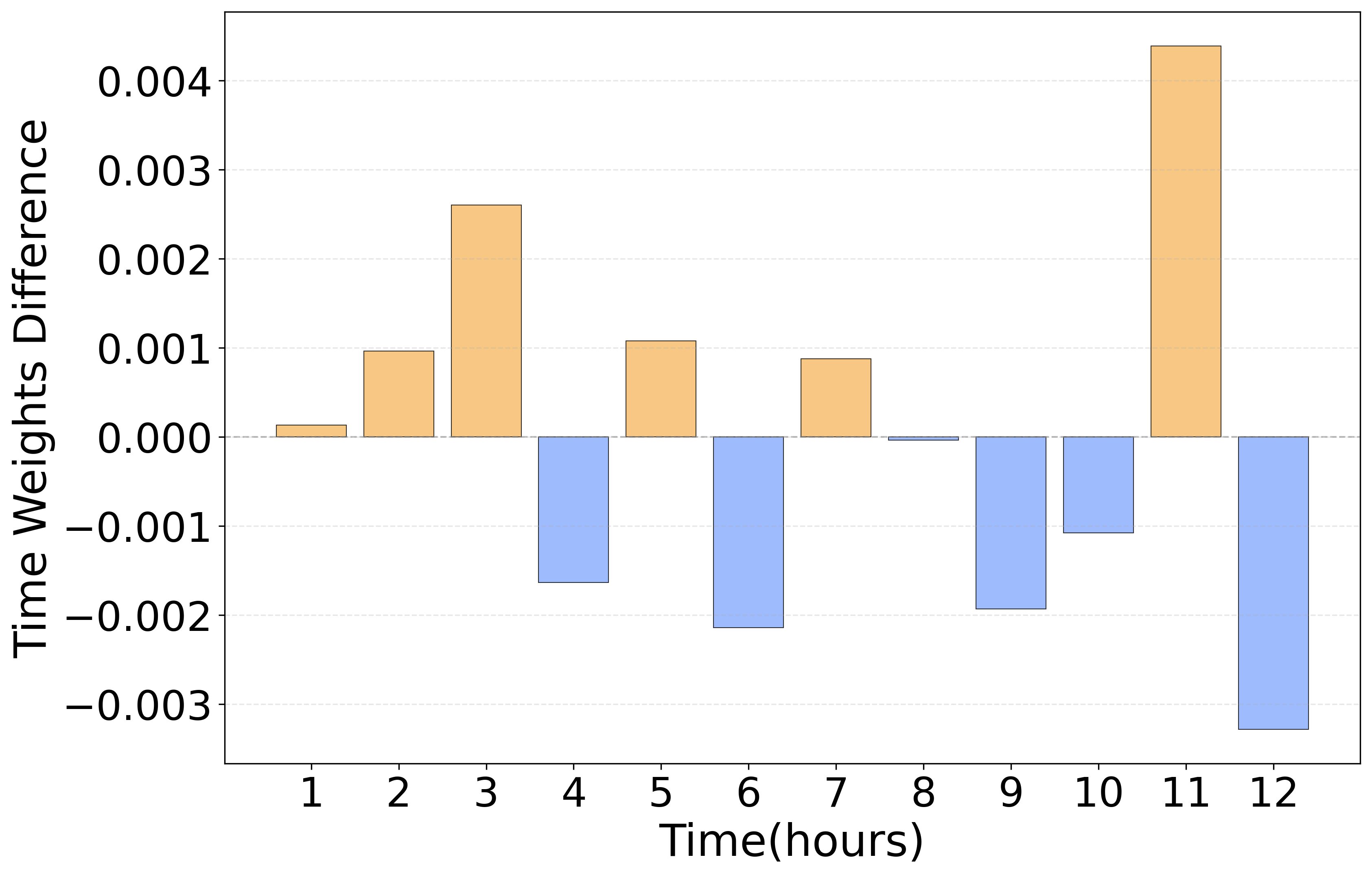}
        \caption{Frost.}
        \label{subfig:2x2e}
    \end{subfigure}
    \begin{subfigure}[b]{0.45\columnwidth}
        \centering
        \includegraphics[width=1.0\textwidth]{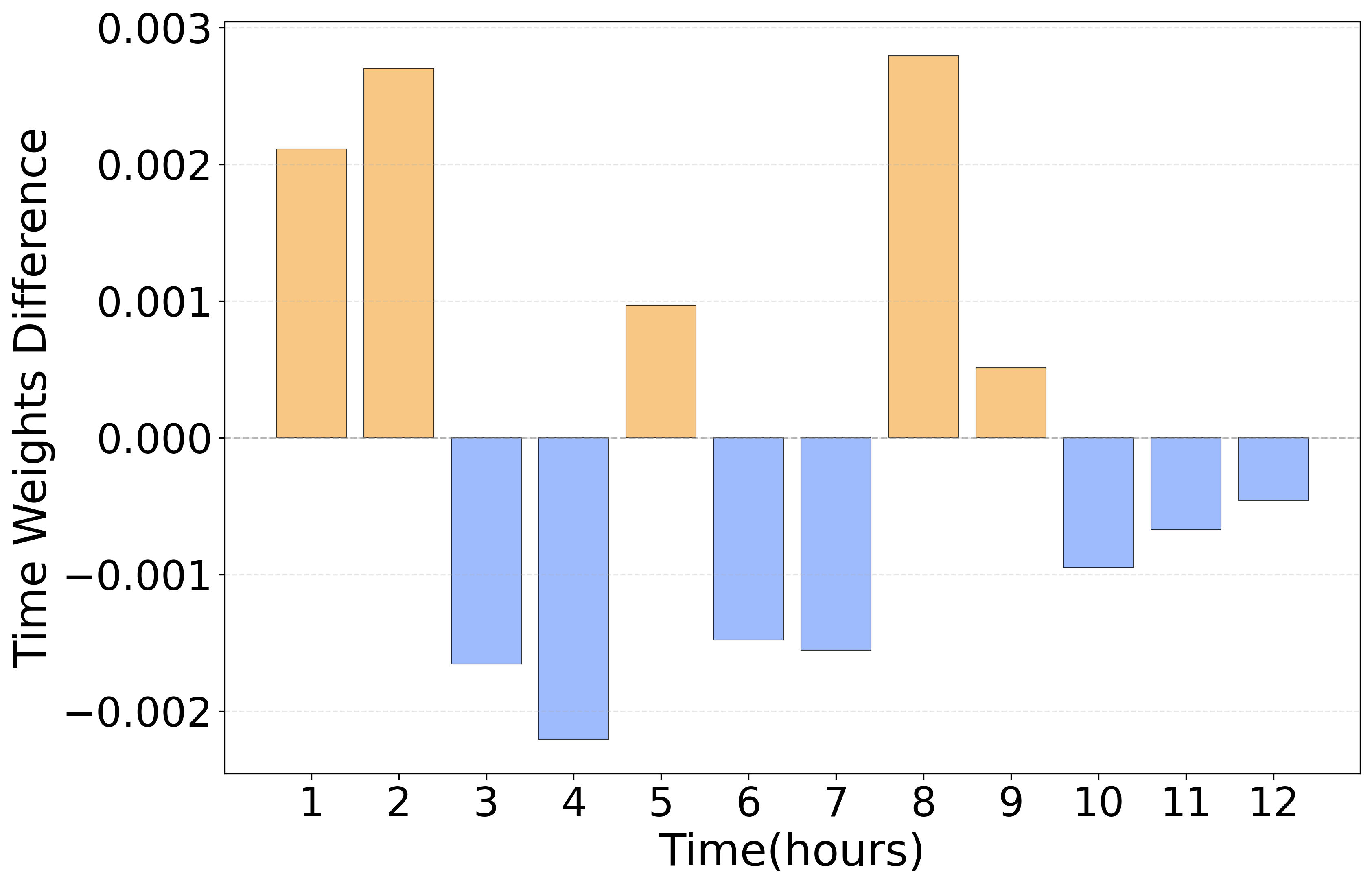}
        \caption{Hail.}
        \label{subfig:2x2f}
    \end{subfigure}
    \caption{Examples of DTGF's weights difference distribution patterns.Each subplot represents a type of severe weather event. The positive bar indicates higher weights for red warnings. The negative bar indicates higher weights for blue warnings.}
    \label{fig:4x2}
\end{figure}

 For snow storms (as shown in \cref{fig:4x2}(b)), DTGF assigns clearly larger positive time-weight differences within the 1–6 h window, while the 9–11 h bins are dominated by negative values. This indicates that severe (red) snowstorm warnings rely more on the rapid intensification during the last 1–6 hours before the event, whereas milder (blue) warnings put relatively higher weights on the earlier 9–12 h evolution. Such a pattern is consistent with the CMA criteria, where red snowstorm warnings are issued when heavy snow ($\geq 15$~mm) is expected within 6 hours, while blue warnings correspond to lighter accumulations ($\geq 4$~mm) over a longer 12-hour window.
 
 For gales (as shown in \cref{fig:4x2}(c)), DTGF produces strong positive time-weight differences in the 1–4 h window, a pronounced negative segment around 7–9 h, and another positive peak near 11 h. This indicates that severe (red) gale warnings rely primarily on the rapid wind strengthening during the last few hours before the event, while milder (blue) warnings place relatively higher weights on the mid-range 7–9 h evolution. The additional positive peak around 11 h suggests that intense gale cases tend to exhibit stronger early precursors than blue-warning cases. Overall, this temporal pattern aligns well with the CMA criteria, where red gale warnings are issued for gales expected within 6 hours, whereas blue warnings are based on the risk of $\geq 6$-grade winds within a much longer 24-hour window.
 
 For hail (as shown in \cref{fig:4x2}(f)), the DTGF module assigns the largest positive time-weight differences to the 1–2 h bins, while the 3–6 h bins are dominated by negative values. This means that severe (red) hail warnings rely much more on the most recent 1–2 h evolution of the storm, whereas milder warnings (e.g., orange) put relatively higher weights on the broader 3–6 h window. This pattern is highly consistent with the CMA operational criteria, where red hail warnings are issued only when hail is highly likely within the next 2 hours, while orange warnings correspond to possible hail within 6 hours.

\section{Scoring Rubric with LLM-as-a-Judge}
In this section, we detail the rubric used to guide GPT-4o’s scoring of NSW answers (see \cref{fig:GPT4o_scoring_criterion}). The rubric asks GPT-4o, acting as a meteorological researcher, to compare the model’s output with the reference warnings and assign a score from 0 to 5. Scores are determined primarily by whether the locations, levels (blue, yellow, orange, red), and categories (eight types: Rain Storm, Snow Storm, Gale, Frost, Cold Wave, Heat Wave, Hail, Normal) match those in the reference. A score of 5 corresponds to completely accurate predictions (all locations and all level–category pairs match), while intermediate scores (4–2) reflect varying degrees of partial coverage or minor/more substantial omissions and mismatches. cores of 1 and 0 are reserved for mostly incorrect or completely irrelevant answers, where only a small fraction—or none—of the reference warnings are correctly reproduced.

\section{Case Study}
\begin{figure*}[!htbp]
    \centering
    \includegraphics[width=1.0\linewidth]{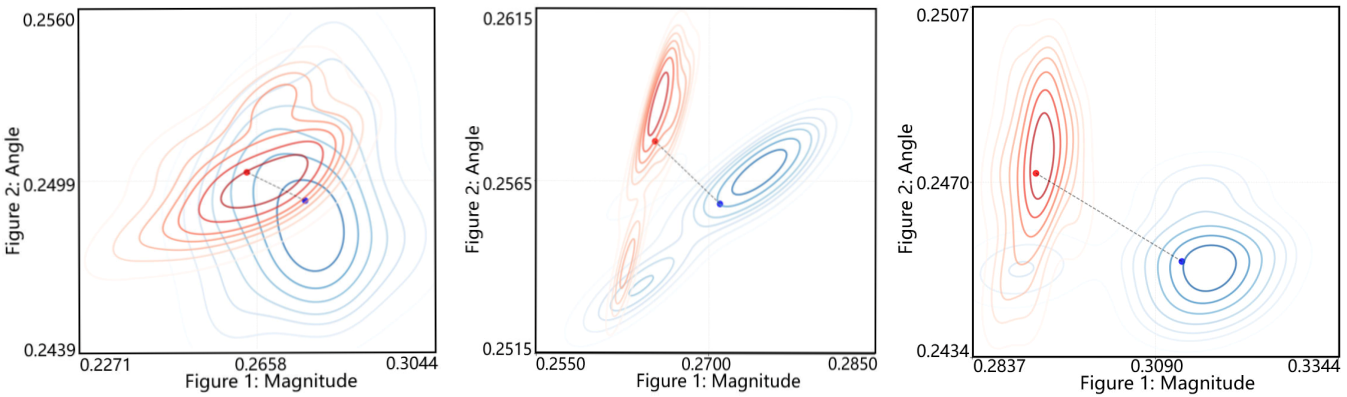}
    \caption{CoE Feature distribution of correct and incorrect sample sets in three Question types: (a) MC questions, (b) T/F questions, (c) RSW questions. Blue and Red distributions represent the correct and incorrect samples. Dataset used in this figure is testing set and model used in the figure is Qwen2.5-VL-7B-Instruct. }
    \label{fig:coe}
\end{figure*}

To better understand how the model attends to meteorological patterns when it succeeds or fails, we compared 20 rainstorm events with correct and incorrect predictions and visualized their COE fields. \cref{fig:coe} shows contour plots of correct samples (blue) versus incorrect samples (red) across the two attention-feature dimensions, namely magnitude and angle, for MC, T/F and RSW tasks. The peak of the incorrect-sample distribution is shifted toward higher magnitude and angle values compared to the correct-sample distribution, and its contours are more dispersed, indicating that the meteorological features the model focuses on when it makes errors are more complex and variable. In particular, many misclassified cases correspond to compound weather situations (e.g., heavy rainfall accompanied by strong winds), where the model tends to overemphasize certain signals such as wind and incorrectly predicts gale events, even when the ground-truth label is a rainstorm.

\vspace{-0.3cm}
\begin{figure*}[!htbp]
\centering
\begin{tcolorbox}[title=True/False Question,fonttitle=\bfseries\LARGE,colback=gray!5,colframe=gray!75!gray,width=1.0\linewidth]
{\Large


\vspace{0.1cm}

As a professional meteorologist, please analyze the provided ERA5 dataset and determine whether \textbf{Maqu County (Coordinates: [34.00°N, 102.07°E])} is currently experiencing severe weather. Respond with either \textbf{"Yes"} or \text{"No"}.

}
\end{tcolorbox}
\vspace{-0.2cm}
\caption{An example of True/False question prompt for training and testing.}
\label{fig:tf_question}
\end{figure*}

\begin{figure*}[!htbp]
\centering
\captionsetup{skip=4pt} 
\begin{tcolorbox}[title=Regional Severe Weather Question,fonttitle=\bfseries\LARGE,colback=gray!5,colframe=gray!75!gray,width=1.0\linewidth]
{\Large


\vspace{0.1cm}

As a professional meteorologist, please analyze the provided ERA5 data and assess the likelihood of severe weather events occurring in \textbf{Fuzhou City (Coordinates: [26.08°N, 119.30°E])}.
Please identify which types of severe weather events may occur, selecting from the following categories:

\vspace{0.2cm}

Rain Storm, Snow Storm, Gale, Cold Wave, Heat Wave, Frost, Hail.

}
\end{tcolorbox}
\caption{An example of RSW question prompt for training and testing.}
\label{fig:rsw_question}
\end{figure*}

\begin{figure*}[!htbp]
\centering
\begin{tcolorbox}[title=Multiple Choice Question,fonttitle=\bfseries\LARGE,colback=gray!5,colframe=gray!75!gray,width=1.0\linewidth]
{\Large


\vspace{0.1cm}

As a professional meteorologist, please identify the severe weather events that occurred in \textbf{Beijing (Coordinates:[39.90°N,116.40°E])} based on the input data. Please select only one applicable option from the following options:

\vspace{0.2cm}

\textbf{[Rain Storm]}

A1: Blue-level ~~~~~~~~~
A2: Yellow-level ~~~~~~~~~
A3: Orange-level ~~~~~~~~~
A4: Red-level

\vspace{0.1cm}

\textbf{[Snow Storm]}

B1: Blue-level ~~~~~~~~~
B2: Yellow-level ~~~~~~~~~
B3: Orange-level ~~~~~~~~~
B4: Red-level

\vspace{0.1cm}

\textbf{[Gale]}

C1: Blue-level ~~~~~~~~~
C2: Yellow-level ~~~~~~~~~
C3: Orange-level ~~~~~~~~~
C4: Red-level

\vspace{0.1cm}

\textbf{[Cold Wave]}

D1: Blue-level ~~~~~~~~~
D2: Yellow-level ~~~~~~~~~
D3: Orange-level ~~~~~~~~~
D4: Red-level

\vspace{0.1cm}

\textbf{[Heat Wave]}

E1: Yellow-level~~~~~~
E2: Orange-level ~~~~~~~~~
E3: Red-level

\vspace{0.1cm}

\textbf{[Frost]}

F1: Blue-level~~~~~~~~~~
F2: Yellow-level ~~~~~~~~~~ 
F3: Orange-level

\vspace{0.1cm}

\textbf{[Hail]}

G1: Orange-level~~~~~
G2: Red-level

\vspace{0.1cm}

\textbf{[Normal Conditions]}

H1: No warnings issued

}
\end{tcolorbox}
\vspace{-0.2cm}
\caption{An example of multiple choice question prompt for training and testing.}
\label{fig:answer_extraction}
\end{figure*}

\begin{figure*}[!htbp]
\centering
\captionsetup{skip=4pt} 
\begin{tcolorbox}[title=National Severe Weather Question,fonttitle=\bfseries\LARGE,colback=gray!5,colframe=gray!75!gray,width=1.0\linewidth]
{\Large


\vspace{0.1cm}

As a professional meteorologist, you are tasked with analyzing the provided dataset to identify and characterize any severe weather events that have occurred across \textbf{China's} administrative divisions. 

Please focus on the following regions with their respective coordinates:

\vspace{0.2cm}

Hebei(Coordinates:[38.04°N,114.51°E]),\\ Shanxi(Coordinates:[37.87°N,112.55°E]), \\
Liaoning(Coordinates:[41.80°N,123.50°E]), \\ Jilin(Coordinates:[43.90°N,125.33°E]), \\
Heilongjiang(Coordinates:[45.76°N,126.64°E]),……

\vspace{0.2cm}

Determine what kind of severe weather occurred in each region. Output only the area where severe weather occurs For each detected event, use the following structured format: [Region Name] issues a [Event Type] [Severity Level]. 

\vspace{0.2cm}

Definitions: 

1. Region Name (Administrative Divisions) 

2. Event Type (Rain Storm/Snow Storm/Gale/Cold Wave/Heat Wave/Frost/Hail)

3. Severity Level (Blue/Yellow/Orange/Red) 

}
\end{tcolorbox}
\caption{An example of NSW question prompt for training and testing.}
\label{fig:nsw_question}
\end{figure*}

\begin{figure*}[!htbp]
\centering
\begin{tcolorbox}[title=GPT-4o's scoring criterion,fonttitle=\bfseries\LARGE,colback=gray!5,colframe=gray!75!gray,width=1.0\linewidth]
{\Large


\vspace{0.1cm}

You are a meteorological researcher. Your task is to compare the reference (ground-truth) answers with the outputs from a multi-modal meteorological model, assign a score from 0 to 5, and explain your reasoning. Each answer consists of multiple warning entries separated by commas, each formatted as: \textless Location\textgreater\textless Level\textgreater\textless Category\textgreater. Categories (8 types): Rain Storm, Snow Storm, Gale, Frost, Cold Wave, Heat Wave, Fog No Severe Weather. Levels (4 colors): Blue, Yellow, Orange, Red. Use the following open-ended scoring rubric (0-5) and explain why you chose that score:

\vspace{0.2cm}

\noindent\textbf{5 -- Completely accurate: }\\[2pt]
\hspace*{1em}\textbullet\; All locations match exactly.\\
\hspace*{1em}\textbullet\; For each warning, Level and Category both match perfectly.

\noindent\textbf{4 -- Accurate but slightly incomplete: }\\[2pt]
\hspace*{1em}\textbullet\; Locations match exactly.\\
\hspace*{1em}\textbullet\; Categories match perfectly; Levels partially match.

\noindent\textbf{3 -- Partially accurate: }\\[2pt]
\hspace*{1em}\textbullet\; Most locations match, with minor omissions or discrepancies.\\
\hspace*{1em}\textbullet\; Categories partially match; Levels partially match.

\noindent\textbf{2 -- Contains some errors: }\\[2pt]
\hspace*{1em}\textbullet\; Some reference locations are missing in the model output.\\
\hspace*{1em}\textbullet\; Categories partially match; Levels partially match.

\noindent\textbf{1 -- Mostly incorrect: }\\[2pt]
\hspace*{1em}\textbullet\; Most reference locations are missing or incorrect.\\
\hspace*{1em}\textbullet\; Only a small fraction of entries match in Location, Level, or Category.

\noindent\textbf{0 -- Completely incorrect or irrelevant: }\\[2pt]
\hspace*{1em}\textbullet\; No locations match.\\
\hspace*{1em}\textbullet\; No entries match in Level or Category.

\vspace{0.2cm}

\textbf{Please score the following:}

\textbf{Standard answer:} [Insert standard answer here]

\textbf{Multimodal meteorological foundation model:} [Insert student's answer here]

Please score the model's output according to the criteria and explain the reasoning for the score given.

}
\end{tcolorbox}
\vspace{-0.2cm}
\caption{GPT-4o's scoring criterion.}
\label{fig:GPT4o_scoring_criterion}
\end{figure*}

\begin{table*}[!htbp]
\centering

\begin{tabular}{lc}
\toprule
\textbf{Parameter} & \textbf{Value} \\
\midrule
\multicolumn{2}{l}{\textbf{Model Parameters}} \\
model\_name\_or\_path & /model\_pre-trained\_weights \\
adapter\_name\_or\_path & - \\
trust\_remote\_code & true \\
\midrule
\multicolumn{2}{l}{\textbf{Method Parameters}} \\
stage & sft \\
do\_train & True \\
finetuning\_type & lora \\
quantization\_method & bitsandbytes \\
template & specific\_model\_templates \\
flash\_attn & auto \\
dataset & train\_data \\
cutoff\_len & 9500 \\
max\_samples & 220000 \\
preprocessing\_num\_workers & 16 \\
\midrule
\multicolumn{2}{l}{\textbf{LoRA Parameters}} \\
lora\_rank & 8 \\
lora\_alpha & 16 \\
lora\_dropout & 0 \\
lora\_target & all \\
additional\_target & gating\_mlp, linear, text\_proj, mlp\_channel \\
\midrule
\multicolumn{2}{l}{\textbf{Training Parameters}} \\
per\_device\_train\_batch\_size & 2 \\
gradient\_accumulation\_steps & 8 \\
learning\_rate & 0.00005 \\
num\_train\_epochs & 3.0 \\
lr\_scheduler\_type & cosine \\
max\_grad\_norm & 1.0 \\
warmup\_steps & 0 \\
packing & False \\
report\_to & none \\
bf16 & True \\
optim & adamw\_torch \\
ddp\_timeout & 180000000 \\
include\_num\_input\_tokens\_seen & True \\
ddp\_find\_unused\_parameters & False \\
seed & 42 \\
\midrule
\multicolumn{2}{l}{\textbf{Output Parameters}} \\
output\_dir & /path/to/output \\
logging\_steps & 8 \\
save\_steps & 200 \\
plot\_loss & True \\
overwrite\_output\_dir & True \\
save\_only\_model & False \\
\bottomrule
\end{tabular}

\caption{Model Training Parameters}
\label{tab:training_params}
\end{table*}

\begin{table*}[!htbp]
\centering

\begin{tabular}{lc}
\toprule
\textbf{Parameter} & \textbf{Value} \\
\midrule
\multicolumn{2}{l}{\textbf{Model Parameters}} \\
model\_name\_or\_path & /model\_pre-trained\_weights \\
adapter\_name\_or\_path & /lora-trained\_weights \\
trust\_remote\_code & true \\
\midrule
\multicolumn{2}{l}{\textbf{Method Parameters}} \\
stage & sft \\
do\_train & False \\
finetuning\_type & lora \\
quantization\_method & bitsandbytes \\
template & specific\_model\_templates \\
flash\_attn & auto \\
dataset\_dir & data \\
eval\_dataset & test\_data \\
cutoff\_len & 9500 \\
max\_samples & 100000 \\
preprocessing\_num\_workers & 16 \\
\midrule
\multicolumn{2}{l}{\textbf{Evaluation Parameters}} \\
per\_device\_eval\_batch\_size & 4 \\
predict\_with\_generate & True \\
max\_new\_tokens & 512 \\
top\_p & 0.7 \\
temperature & 0.95 \\
do\_predict & True \\
seed & 42 \\
\midrule
\multicolumn{2}{l}{\textbf{Output Parameters}} \\
output\_dir & /path/to/output \\
\bottomrule
\end{tabular}

\caption{Model Evaluation Parameters}
\label{tab:evaluation_params}
\end{table*}


\end{document}